\RequirePackage{fix-cm}
\documentclass[twocolumn]{svjour3}          
\smartqed  
\usepackage{graphicx}
\usepackage{xcolor}
\usepackage{tabularx}
\usepackage{multirow}
\usepackage{graphicx}
\usepackage{makecell}
\usepackage{amsmath}
\usepackage{color}
\usepackage{float}
\usepackage{rotating}
\usepackage{placeins}
\usepackage{amsfonts}
\usepackage{bm}

\usepackage[round,authoryear,comma,sort&compress,numbers]{natbib} 
\usepackage[colorlinks=true,linkcolor=blue,citecolor=black,urlcolor=blue]{hyperref}

\usepackage{diagbox}
\usepackage{cancel} 
\usepackage{mathrsfs} 
\usepackage{eqnarray} 
\usepackage{subcaption}

\usepackage{pifont}

\newcommand{\changed}[1]{\textcolor{black}{#1}}

\begin{document}

\title{Surveillance Face Recognition Challenge
}


\author{Zhiyi~Cheng \and
	Xiatian~Zhu \and
	Shaogang~Gong   
}


\institute{Zhiyi Cheng and Shaogang Gong \at School
of Electrical Engineering and Computer Science, Queen Mary University of London, 
London E1 4NS, UK. \\
\email{\{z.cheng, s.gong\}@qmul.ac.uk}
\and
Xiatian Zhu \at Vision Semantics Limited, London E1 4NS, UK. \\
\email{eddy@visionsemantics.com}
}


\date{Received: date / Accepted: date}

\maketitle

\begin{abstract}

Face recognition (FR) is one of the most extensively investigated problems
in computer vision.
Recently we have witnessed significant progress in FR
with the help of deep learning algorithms and larger scale
datasets particularly those {\em constrained} social media web images,
e.g. high-resolution photos of celebrity faces taken by professional photo-journalists. 
However, the more challenging FR in {\em unconstrained} and {\em low-resolution} 
surveillance images
remains open and under-studied.

To stimulate the development of innovative FR methods effective
and robust for low-resolution surveillance facial images,
we introduce a new Surveillance Face Recognition Challenge, namely {\em QMUL-SurvFace}.
This benchmark is the {\em largest} to date and more importantly the only {\em true} surveillance FR benchmark
to our best knowledge. 
The low-resolution facial images are not synthesised
by artificial down-sampling of high-resolution imagery
but drawn from real surveillance videos.
This benchmark has 463,507 facial images of 15,573 unique identities
captured in uncooperative surveillance scenes
over wide space and time.
Consequently, QMUL-SurvFace presents a true-performance surveillance FR challenge
characterised by low-resolution, motion blur, uncontrolled poses, 
varying occlusion,
poor illumination, and background clutters.

On the QMUL-SurvFace challenge,
we benchmark the FR performances of five representative deep learning face
recognition models
(DeepID2, CentreFace, VggFace, FaceNet, and SphereFace),
in comparison to existing benchmarks.
We show that the current state of the arts are still {\em far from being satisfactory}
to tackle the under-investigated surveillance FR problem. In particular, we demonstrate a surprisingly large gap in FR performance between
existing celebrity photoshot benchmarks (e.g. MegaFace) and 
QMUL-SurvFace.
For example, the CentreFace model 
only yields 29.9\% in Rank-1 rate on {QMUL-SurvFace}, 
{\em vs} a much higher performance
at $65.2$\% on MegaFace in a closed-set test.
Besides, open-set FR is shown more challenging but
required for surveillance scenarios due to the presence of a large number
of non-target people (distractors) in public space.
This is indicated by weak FR performances on QMUL-SurvFace, 
e.g. the top-performer 
CentreFace 
achieves {13.8}\% {\em success rate} (at Rank-20)
at the $10\%$ {\em false alarm rate}.

Moreover, we extensively investigate the challenging low-resolution
problem inherent to the surveillance facial imagery by testing
different combining methods of image super-resolution and face recognition models
on both web and surveillance datasets.
Our evaluations suggest that
current super-resolution methods are
ineffective in improving low-resolution FR performance,
although they have been shown useful 
to synthesise high frequency details 
for low-resolution web images.

Finally, we discuss a number of open research directions and problems 
that need to be addressed for solving the surveillance FR challenge.
The QMUL-SurvFace challenge is publicly available at
\url{https://qmul-survface.github.io/}.

\end{abstract}

\begin{figure*} [h]
	\centering
	\includegraphics[width=1\linewidth]{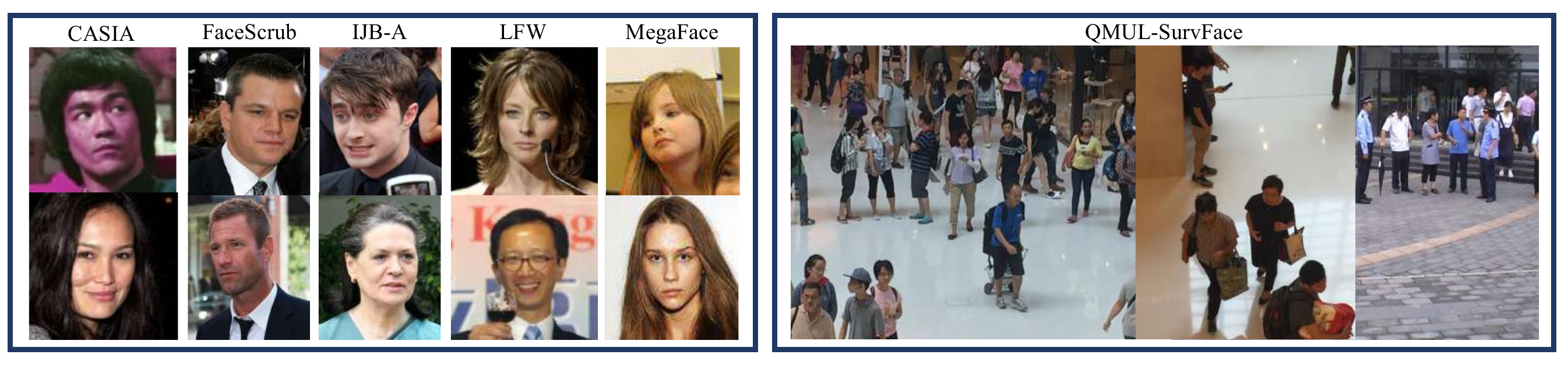}
	\vskip -0.0cm
	\caption{Example comparisons of ({\bf Left}) {\em web face images} from five 
		 standard face recognition challenges and 
		({\bf Right})  {\em native surveillance face images} from typical 
		public surveillance scenes in real-world applications.
	}
	\label{fig:FR_datasets}
\end{figure*}

\section{Introduction}
\label{intro}

Face recognition (FR) is a well established research problem in computer vision with 
the objective to recognise human identities (IDs) by their facial images \citep{zhao2003face}. 
There are other visual recognition based biometrics approaches to
human identification, for example, whole-body person re-identification \citep{gong2014person}, 
iris recognition \citep{phillips2010frvt}, 
gait recognition \citep{sarkar2005humanid}, and
fingerprint recognition \citep{maltoni2009handbook}.
Among these, FR is considered as one of the more convenient
and non-intrusive means for a wide range of identification applications from 
law enforcement and information security to business, entertainment
and e-commerce
\citep{wechsler2012face}. This is due to one fact that
face appearance is more reliable and stable than clothing
appearance, provided that facial images
are visible. 

The FR problem has been extensively studied over the past few decades
since 1970s, and increasingly deployed in social 
applications within the last five years. 
In 2016 alone, there were over 77,300 FR publications including patents by Google Scholar.
%
In particular, the last few years have witnessed FR performance on
high-resolution good quality web images reaching a level arguably better than
the humans, e.g. $99.83\%$ for 1:1 face verification on the LFW challenge 
and \changed{$99.801\%$} for 1:N face identification on the million-scale MegaFace challenge. 
Commercial FR products have become more mature and
appeared increasingly in our daily life, e.g. web photo-album, online e-payment,
and smart-phone e-banking, with such a trend 
accelerating at larger scales. 
One driving force behind the FR success is the 
synergistic and rapid advances in
neural network based deep learning techniques, 
large facial imagery benchmarks, 
and powerful computing devices. 
The progress is advancing rapidly under the joint efforts of both the academic and industrial communities. 
%
%
One may argue then: 
``{\em The state-of-the-art FR algorithms, especially with the help of
  deep learning on large scale data, have reached a sufficient level of maturity
	and application readiness, evidenced by almost saturated
        performances on large scale public benchmark challenges\footnote{The FR performance differences among top-performing models
	are very marginal and negligible (see Table
        \ref{table:LFW}).}. Therefore, the FR problem should be considered as
        ``solved'' and the remaining efforts lie mainly in system
	production engineering.}''

However, as shown in this study, current FR methods scale poorly to
natively noisy and low-resolution images (not synthesised by
down-sampling), that are typified by facial 
data captured in {\em unconstrained wide-field surveillance
  videos and images},
perhaps one of the most important FR application fields in practice.
Specifically, we demonstrate that FR in the typical surveillance images is far away 
from being satisfactory especially at a large scale.
Unlike recognising high-resolution web photoshot images with limited
noise, the problem of surveillance FR remains extremely
challenging and open.
%
This is a surprise because surveillance video data are
characterised by low-resolution imagery with heavy noise,
subject to poor imaging conditions giving rise to
unconstrained pose, expression, occlusion, lighting and background clutter
(Fig. \ref{fig:FR_datasets}).

In the literature, the {\em surveillance face recognition} problem is 
significantly under-studied in comparison to FR on web
photoshots. Whilst {\em web face recognition} is popular and
commercially attractive due to
the proliferation of social media and e-commerce on smart-phones
and various digital devices, 
solving the surveillance FR challenge is critical for
public safety and law enforcement applications.
A major reason for lacking the development of robust and scalable FR
models suitable for surveillance scenes is the lack of large
scale true surveillance facial imagery benchmark,
in contrast to the rich 
availability of high-resolution web photoshot FR benchmarks
(Table \ref{table:existDatasets}).
For example, there are 4,753,320 web face images from 
672,057 face IDs in the MegaFace2
benchmark\footnote{\url{http://megaface.cs.washington.edu/}} \citep{nech2017level}.
This is made possible by easier collection and
labelling of large scale facial images in the public domain 
from the Internet.
%

On the contrary, it is prohibitively expensive and less feasible
to construct large scale {\em native} surveillance facial imagery data as a
benchmark for wider studies, due both to largely restricted data access
and very tedious data labelling at high costs.
Currently, the largest surveillance FR challenge benchmark is 
the UnConstrained College Students (UCCS)
dataset\footnote{\url{http://vast.uccs.edu/Opensetface/}}
\citep{gunther2017unconstrained}, which contains
100,000 face images from 1,732 face IDs,
at a significantly smaller scale than the MegaFace celebrities
photoshot dataset.
However, the UCCS is limited and only semi-native 
due to being captured by a high-resolution camera
at a single location. 
In this study, we show that: 
(1)
The state-of-the-art deep learning based FR models trained on large
scale high-quality benchmark datasets such as the MegaFace
generalise poorly to native low-quality surveillance face recognition tasks; 
(2) The FR
performance test on artificially synthesised low-resolution
images 
does not well reflect the true challenges of
native surveillance FR 
in system deployments;
\changed{(3) 
The image super-resolution models suffer the lack of high-resolution
surveillance facial images which are necessary for model training,
apart from the domain distribution
shift between web and surveillance data.}

In this study, we provide a more realistic and large
scale {\em Surveillance Face Recognition Challenge} than what is
currently available in the public domains.  That is,
recognising a person's identity by natively low-resolution
surveillance facial images taken from unconstrained public scenes.
We make three {contributions} as follows:\\
{\bf ({I})}
We construct a dataset of large scale face IDs with {\em
  native} surveillance facial imagery data for
model development and evaluation. 
Specifically, we
introduce the {\em QMUL-SurvFace} challenge,
containing 463,507 face images of 15,573
IDs. 
%
To our best knowledge, this is the largest sized dataset for
surveillance FR challenge, with {\em native}
low-resolution surveillance facial images captured by unconstrained 
wide-field cameras at distances.
This benchmark dataset is constructed by data-mining $17$ public domain person 
re-identification datasets
(Table \ref{table:reid_dataset}) 
using a deep learning face detection model, so to assemble a large pool of
labelled surveillance face images 
in a {\em cross-problem} data re-purposing principle. 
%
A unique feature of this new surveillance FR benchmark
is 
the provision of cross-location (cross camera views) ID label annotations, available from
the source person re-identification datasets. This cross-view
labelling information is useful for open-set 
FR test. \\
%
%
{\bf ({II})}
We benchmark five representative deep learning FR models \citep{sun2014deepid2,wen2016discriminative,schroff2015facenet,parkhi2015deep,liu2017sphereface}
for both face identification and verification.
In contrast to existing FR challenges typically considering a {\em closed-set} setting,
we particularly evaluate these algorithms in performing a 
more realistic {\em open-set} surveillance FR task,
originally missing in the previous studies.
The closed-set test assumes the existence of every probe face ID
in the gallery set so a true-match always exists for each probe, whilst
the open-set test does not, 
respecting realistic large surveillance
FR scenarios. \\
{\bf ({III})} 
We investigate the effectiveness of existing FR models on native
low-resolution surveillance imagery data by exploiting simultaneously 
image super-resolution
\citep{dong2014learning,dong2016accelerating,LapSRN,kim2016accurate,DRRN17}
and FR models.
%
We study different combinations of super-resolution and FR 
models including independent and joint training schemes.
%
%
We further compare model performances on
MegaFace and UCCS benchmarks to give better understanding of the
unique characteristics of QMUL-SurvFace.
We finally provide extensive discussions on future research directions
towards solving the surveillance FR challenge. 

\begin{table*}
	\centering
	\setlength{\tabcolsep}{6pt}
	\caption{
		The statistics of representative publicly available face recognition 
		benchmarks. 
%
		Celeb: Celebrity.
	}
	\vskip -0.2cm
	\label{table:existDatasets}
	\begin{tabular}{l||c|c|c|c|c|c}
		\hline 
		Challenge 
		& Year & IDs & Images & Videos  & Subject & Surveillance? \\
		\hline 
		Yale \citep{belhumeur1997eigenfaces}
		& 1997
		& 15 & 165 & 0 & Cooperative & No 
		\\
		
		\color{black}QMUL-MultiView \citep{gong2000}
		& \color{black}1998
		& \color{black}25 & \color{black}4,450 & \color{black}5 & \color{black}Cooperative & \color{black}No 
		\\
		
		XM2VTS \citep{messer1999xm2vtsdb}
		& 1999
		& 295 & 0 & 1,180 & Cooperative & No 
		\\
		
		Yale B \citep{georghiades2001few} 
		& 2001 
		& 10 & 5,760 & 0 & Cooperative & No 
		\\
		
		CMU PIE \citep{sim2002cmu} 
		& 2002
		& 68 & 41,368 & 0 & Cooperative & No 
		\\
		
		Multi-PIE \citep{gross2010multi} 
		& 2010
		& 337 & 750,000 & 0 & Cooperative & No 
		\\ 
		\hline
		
		Morph \citep{ricanek2006morph} 
		& 2006
		& 13,618 & 55,134 & 0 & Celeb (Web) & No 
		\\ 
			
		LFW \citep{huang2007labeled}   
		& 2007
		& 5,749 &  13,233  & 0 & Celeb (Web) & No 
		\\ 
		
		YouTube \cite {wolf2011face} 
		& 2011
		&   1,595   &  0  & 3,425 & Celeb (Web) & No 
		\\
		
		WDRef \citep{chen2012bayesian} 
		& 2012 
		& 2,995 & 99,773 & 0  & Celeb (Web) & No 
		\\ 
		
		FaceScrub \citep{ng2014data} 
		& 2014
		& 530 & 100,000 & 0 & Celeb (Web) & No 
		\\ 
		
		CASIA \citep{yi2014learning} 
		& 2014
		& 10,575 & 494,414 & 0  & Celeb (Web) & No 
		\\ 
		
		CelebFaces \citep{celebface} 
		& 2014
		& 10,177 & 202,599 & 0  & Celeb (Web) & No 
		\\ 
		
		IJB-A \citep{klare2015pushing} 
		& 2015
		& 500  & 5,712   & 2,085 & Celeb (Web) & No 
		\\ 

		VGGFace \citep{parkhi2015deep}  
		& 2015
		& 2,622 & 2.6M  & 0 & Celeb (Web) & No 
		\\

		UMDFaces \citep{bansal2016umdfaces}
		& 2016
		& 8,277 & 367,888 & 0  & Celeb (Web) & No 
		\\
		
		CFP~\citep{sengupta2016frontal} 
		& 2016 
		& 500 & 7,000 & 0 & Celeb (Web) & No 
		\\ 
		
		UMDFaces \citep{bansal2016umdfaces} 
		& 2016
		& 8,277 & 367,888 & 0 & Celeb (Web) & No 
		\\ 
		
		MS-Celeb-1M \citep{guo2016ms} 
		& 2016
		& 99,892 & 8,456,240 & 0 & Celeb (Web) & No 
		\\ 
		
		UMDFaces-Videos \citep{bansal2017s} 
		& 2017
		& 3,107 & 0 & 22,075 & Celeb (Web) & No 
		\\
		
		IJB-B \citep{whitelam2017iarpa} 
		& 2017
		& 1,845  & 11,754   & 7,011 & Celeb (Web) & No 
		\\
		
		VGGFace2 \citep{cao2017vggface2} 
		& 2017 
		& 9,131 & 3.31M & 0 & Celeb (Web) & No
		\\
		
		MegaFace2 \citep{nech2017level} 
		& 2017
		& 672,057 & 4,753,320 & 0 & Non-Celeb (Web) & No 
		\\ 
		
		\hline 

		FERET \citep{phillips2000feret} 
		& 1996 
		& 1,199 
		& 14,126 & 0 & Cooperative & No 
		\\
		
		FRGC \citep{phillips2005overview} 
		& 2004 
		& 
		466+ 
		& 50,000+
		& 0 & Cooperative & No 
		\\
		
		{\color{black} CAS-PEAL \citep{gao2008cas} }
		& {\color{black} 2008}
		& 
		{\color{black} 1,040}
		& 
		{\color{black} 99,594} 
		& {\color{black} 0} & {\color{black} Cooperative} & {\color{black} No} 
		\\
		
		PaSC \citep{beveridge2013challenge} 
		& 2013 & 293 & 9,376 & 2,802 & Cooperative & No 
		\\

		FRVT(Visa) \citep{ngan2017face} 
		& 2017
		& O($10^5$) & O($10^5$)
		& 0 & Cooperative & No 
		\\
		FRVT(Mugshot) \citep{ngan2017face} 
		& 2017
		& O($10^5$) & O($10^6$)
		& 0 & Cooperative & No 
		\\
		FRVT(Selfie) \citep{ngan2017face} 
		& 2017
		& $<$500 & $<$500
		& 0 & Cooperative & No 
		\\
		FRVT(Webcam) \citep{ngan2017face} 
		& 2017
		& $<$1,500 & $<$1,500
		& 0 & Cooperative & No 
		\\
		FRVT(Wild) \citep{ngan2017face} 
		& 2017
		& O($10^3$) & O($10^5$)
		& 0 & Uncooperative & No
		 \\
		FRVT(Child Exp) \citep{ngan2017face} 
		& 2017
		& O($10^3$) & O($10^4$)
		& 0 & Uncooperative  & No 
		\\

		
		
		\hline
		SCface~\citep{grgic2011scface} 
		& 2011
		& 130 
		& 4,160 & 0 & Cooperative & Yes 
		\\

		{\color{black} COX \citep{huang2015benchmark} }
		& {\color{black} 2015}
		& 
		{\color{black} 1,000}
		& 
		{\color{black} 1,000} 
		& {\color{black} 3,000} & {\color{black} Cooperative} & {\color{black} Yes} 
		\\
		
		
		UCCS \citep{gunther2017unconstrained} 
		& 2017
		& 1,732 & 14,016+
		& 0 & Uncooperative & Yes 
		\\
		%
		
		\textbf{QMUL-SurvFace}
		& 2018
		& {15,573} & {463,507} & 0 & Uncooperative & Yes 
		\\
		\hline 
	\end{tabular}
\end{table*}

\section{Related Work}\label{sec:existMethods}
In this section, we review and discuss the representative 
FR challenges (Sec. \ref{sec:FRDatasets})
and methods (Sec. \ref{sec:ext_FR_methods})
in the literature.
In Sec. \ref{sec:ext_FR_methods}, we focus on the models closely related
to the surveillance FR challenge including recent deep learning  
algorithms. More general and extensive reviews can be found in 
other surveys \citep{prado2016automatic,wang2014low,ersotelos2008building,zou2007illumination,
	abate20072d,tan2006face,kong2005recent,chang2005evaluation,zhao2003face,
	adini1997face,daugman1997face,chellappa1995human,samal1992automatic} 
and books \citep{wechsler2012face,wechsler2009reliable,zhao2011face,zhou2006unconstrained,gong2000,wechsler1998,li2011}.


\subsection{Face Recognition Challenges}\label{sec:FRDatasets}
An overview of representative FR challenges and benchmarks are summarised in Table \ref{table:existDatasets}.
Specifically, early challenges focus on {\em small-scale} {\em constrained} FR scenarios 
with limited number of images and IDs
\citep{samaria1994parameterisation,belhumeur1997eigenfaces,messer1999xm2vtsdb,georghiades2001few,sim2002cmu,gross2010multi,phillips2010frvt}.
They provide neither sufficient appearance variation and diversity for
robust model training, nor practically solid test benchmarks.
In 2007, the seminal Labeled Faces in the Wild (LFW) challenge \citep{huang2007labeled} 
was proposed and started to shift the community
towards recognising unconstrained celebrity faces by providing
web face images and a standard performance evaluation protocol.
LFW has contributed significantly to a spurring of interest and progress
in FR.
This trend towards large datasets has been amplified
by the creation of even larger FR benchmarks
such as CASIA \citep{yi2014learning}, CelebFaces \citep{celebface}, VGGFace \citep{parkhi2015deep},
MS-Celeb-1M \citep{guo2016ms}, MegaFace \citep{kemelmacher2016megaface}
and  MegaFace2 \citep{nech2017level}.
Thus far, it seems that the availability of large training and
test data benchmark for web photoshot images has been addressed. 

With such large benchmark challenges,
FR accuracy in good quality images has
reached an unprecedented level by deep learning. 
For example, the FR performance has reached 99.83\% (face verification) on LFW 
and \changed{$99.801\%$} (face identification) on MegaFace. 
However, 
this dose not
scale to {\em native} low-resolution surveillance facial
imagery data captured in unconstrained camera views 
(see Sec. \ref{sec:eval_LR_FR}).
This is due to two reasons:
(1) Existing FR challenges such as LFW have varying degrees of 
data selection bias (near-frontal pose, less motion blur, good
illumination); and 
(2) Deep learning methods are often domain-specific (i.e. 
only generalise well to face images similar to the training data)
On the other hand, there
is big difference in facial images between a web
photoshot view and a surveillance view in-the-wild (Fig. \ref{fig:FR_datasets}).

Research on surveillance FR has made little progress
since the early days in 1996 when the well-known FERET challenge was
launched \citep{phillips2000feret}. It is under-studied by
large, with a very few benchmarks available.
One of the major obstacles is the difficulty of establishing a large
scale surveillance FR challenge due to the high cost and limited feasibility
in collecting surveillance facial imagery data and
exhaustive facial ID annotation.
Even in the FERET dataset, only simulated (framed) surveillance face
images were collected in most cases with carefully controlled imaging
settings, therefore it provides a much better facial image quality
than those from native surveillance videos.

A notable recent study introduced the UCCS face challenge
\citep{gunther2017unconstrained}, which is currently the largest
surveillance FR benchmark in the public domain. 
The UCCS facial images were captured from a
long-range distance without subjects' cooperation (unconstrained).
The faces in these images are of various poses, blurriness and occlusion 
(Fig. \ref{fig:UCCS-QMUL}(b)).
This benchmark represents a relatively realistic surveillance FR scenario in
comparison to FERET. 
However, the UCCS images were captured at high-resolution from a
single camera view\footnote{A
  single Canon 7D camera equipped with a Sigma 800mm F5.6 EX APO DG HSM
  lens.}, therefore providing significantly more facial details with less
viewing angle variations.
Moreover, UCCS is small in size, particularly in term
of the face ID numbers (1,732), statistically
limited for evaluating surveillance FR
challenge (Sec. \ref{sec:eval_LR_FR}). 
In this study, we address the limitations of UCCS
by constructing a larger scale natively low-resolution surveillance
FR challenge, the QMUL-SurvFace benchmark. 
It consists of 463,507 real-world surveillance face images of 15,573 different IDs 
captured from a diverse source of public spaces (Sec. \ref{sec:QMUL_SurvFace}).

\subsection{Face Recognition Methods} \label{sec:ext_FR_methods}
We provide a brief review on the vast number of existing FR algorithms,
including hand-crafted, deep learning, and low-resolution methods. 
The low-resolution FR methods are included 
for discussing the state-of-the-art in handling 
poor image quality. 
We also discuss image super-resolution (hallucination)
techniques for enhancing image fidelity and improving FR
performance on low-resolution imagery data.

\begin{table*} 
	\centering
	\caption{Face verification performance of state-of-the-art FR methods on 
		the LFW challenge. 
		``$^*$'': Results from the challenge leaderboard \citep{LFW}.
		M: Million.
	}
	\setlength{\tabcolsep}{0.2cm}
	\vskip -0.2cm
	\label{table:LFW}
	\begin{tabular}{c|l|c|c|c|c}
		\hline
		Feature Representation & Method & Accuracy (\%) & Year & Training IDs & Training Images \\
		\hline \hline
		\multirow{4}{*}{Hand Crafted} 
		& Joint-Bayes~\citep{chen2012bayesian} & 92.42 & 2012 & 2,995 & 99,773  \\
		& HD-LBP~\citep{chen2013blessing} & 95.17 & 2013 & 2,995 & 99,773  \\
		& TL Joint-Bayes~\citep{cao2013practical} & 96.33 & 2013 & 2,995 & 99,773 \\
		& GaussianFace~\citep{lu2015surpassing} & 98.52 & 2015 & 16,598 & 845,000 \\ 
		\hline
		\multirow{14}{*}{Deep Learning}
		& DeepFace~\citep{taigman2014deepface} & 97.35 & 2014 & 4,030 & 4.18M \\
		& DeepID~\citep{sun2014deepid} & 97.45 & 2014 & 5,436 & 87,628 \\
		& LfS~\citep{yi2014learning} & 97.73 & 2014 & 10,575 & 494,414 \\
		& Fusion~\citep{taigman2015web} & 98.37 & 2015 & 250,000 & 7.5M \\ 
		& {VggFace~\citep{parkhi2015deep}} & {98.95} & 2015 & 2,622 & 2.6M \\
		& BaiduFace~\citep{liu2015targeting} & 99.13 & 2015 & 18,000 & 1.2M \\ 
		& {DeepID2~\citep{sun2014deepid2}} & {99.15} & 2014 & 10,177 & 202,599 \\
		& {CentreFace~\citep{wen2016discriminative}} & {99.28} & 2016 & 17,189 & 0.7M \\ 
		& {SphereFace~\citep{liu2017sphereface}} & {99.42} & 2017 & 10,575 & 494,414 \\ 
		& DeepID2+~\citep{sun2015deeply} & 99.47 & 2015 & 12,000 & 290,000 \\
		& {DeepID3~\citep{sun2015deepid3}} & {99.53} & 2015 & 12,000 & 290,000 \\
		& {FaceNet~\citep{schroff2015facenet}} & {99.63} & 2015 & 8M & 200M\\
		& {TencentYouTu$^*$~\citep{Tencent}} & {99.80} & 2017 & 20,000 & 2M \\ 
		& {EasenElectron$^*$~\citep{EasenElectron}} & {99.83} & 2017 & 59,000 & 3.1M \\ 
		\hline
	\end{tabular}
\end{table*}

\vspace{0.1cm}
\noindent {\bf (I) Hand Crafted Methods.}
Most early FR methods rely on hand-crafted features (e.g. Color Histogram, LBP, SIFT, Gabor) 
and matching model learning algorithms 
(e.g. discriminative margin mining, subspace learning, dictionary based sparse coding, Bayesian modelling)
\citep{turk1991eigenfaces,belhumeur1997eigenfaces,chen2013blessing,ahonen2006face,tan2010enhanced,cao2013practical,zhang2007histogram,liu2002gabor,wright2009robust,chen2012bayesian,wolf2013svm}.
These methods are often inefficient subject to heavy computational cost
of high-dimensional feature representations and complex image preprocessing.
Moreover, they also
suffer from sub-optimal recognition generalisation 
particularly in a large sized data when there are significant
facial appearance variations. 
This is jointly due to weak representation power (human domain knowledge used in hand-crafting features is limited and incomplete)
and 
the lack of end-to-end interaction learning between feature extraction and model inference.

\begin{table*} 
	\centering
	\setlength{\tabcolsep}{0.42cm}
	\caption{
		The performance summary of state-of-the-art image super-resolution methods on five 
		standard benchmarks. 
		While none of the benchmarks are designed for FR,
		these evaluations represent a generic capability of contemporary super-resolution techniques
		in synthesising high frequency details particularly for low-resolution web images.
		Metric: Peak Signal-to-Noise Ratio (PSNR), higher is better.}
	\vskip -0.2cm
	\label{table:srperformance}
	\begin{tabular}{l|c|ccccc}
		\cline{1-7}
		Model & Unscalling Times 
		& Set5 
		& Set14 
		& B100 
		& URGAN 
		& MANGA 
		\\
		\hline \hline 
		Bicubic & 2 & 33.65 & 30.34 &  29.56 & 26.88  & 30.84 \\
		SRCNN~\citep{dong2014learning} & 2 &  36.65 & 32.29 & 31.36 & 29.52 & 35.72 \\
		FSRCNN~\citep{dong2016accelerating} & 2 & 36.99 & 32.73 & 31.51 & 29.87 & 36.62 \\
		VDSR~\citep{kim2016accurate} & 2 & 37.53 & 33.03 & 31.90 & 30.76 & - \\
		DRCN~\citep{kim2016deeply} & 2 & 37.63 & 32.98 & 31.85 & 30.76 & 37.57 \\ 
		LapSRN~\citep{LapSRN} & 2 &  37.52 & 33.08 & 31.80 & 30.41 & 37.27 \\
		DRRN~\citep{DRRN17} & 2 & 37.74 & 33.23 & 32.05 & 31.23 & - \\
		\cline{1-7}
		Bicubic & 4 &  28.42 & 26.10 & 25.96 & 23.15& 24.92 \\
		SRCNN~\citep{dong2014learning} & 4 &  30.49 & 27.61 & 26.91 & 24.53 & 27.66  \\
		FSRCNN~\citep{dong2016accelerating} & 4 & 30.71 & 27.70 & 26.97 & 24.61 & 27.89  \\
		VDSR~\citep{kim2016accurate} & 4 & 31.35 & 28.01 & 27.29 & 25.18 & - \\
		DRCN~\citep{kim2016deeply} & 4 & 31.53 & 28.04 & 27.24 & 25.14 & 28.97 \\ 
		LapSRN~\citep{LapSRN} & 4 &  31.54 & 28.19 & 27.32 & 25.21 & 29.09 \\
		DRRN~\citep{DRRN17} & 4 & 31.68 & 28.21 & 27.38 & 25.44 & - \\
		\cline{1-7}
		Bicubic & 8 & 24.39 & 23.19 & 23.67 & 20.74 & 21.47 \\
		SRCNN~\citep{dong2014learning} & 8 &  25.33 & 23.85 & 24.13 &  21.29 & 22.37  \\
		FSRCNN~\citep{dong2016accelerating} & 8 & 25.41 & 23.93 & 24.21 & 21.32 & 22.39   \\
		LapSRN~\citep{LapSRN} & 8 &  26.14 & 24.44 & 24.54 & 21.81 & 23.39 \\
		\cline{1-7}
	\end{tabular}
\end{table*}

\vspace{0.1cm}
\noindent {\bf (II) Deep Learning Methods.}
In the past five years, FR models based on deep learning, 
particularly convolutional neural networks (CNNs)
\citep{sun2014deepid,taigman2014deepface,klare2015pushing,schroff2015facenet,parkhi2015deep,masi2016pose,wen2016discriminative,kemelmacher2016megaface,liu2017sphereface},
have achieved remarkable success (Table \ref{table:LFW}).
This paradigm benefits from superior network architectures 
\citep{krizhevsky2012imagenet,simonyan2014very,szegedy2015going,he2016deep} 
and 
optimisation algorithms \citep{schroff2015facenet,sun2014deepid,wen2016discriminative}.
Deep FR methods naturally address the limitations of hand-crafted alternatives
by jointly learning face representation and matching model end-to-end. 
To achieve good performance, a large set of labelled face images
is usually necessary to train the millions of parameters of deep models.
This can be commonly satisfied by using millions of web face images 
collected and labelled (filtered) from the Internet sources.
%
Consequently, modern FR models are often trained, evaluated and deployed 
on web face datasets (Table \ref{table:existDatasets} and Table \ref{table:LFW}).

While the web FR performance achieves an unprecedented level,
it remains unclear how well the state-of-the-art methods generalise to
poor quality surveillance facial data. 
Intuitively, more challenges are involved in the surveillance FR scenario
due to three reasons:
{\bf (1)} Surveillance face images contain much less appearance details with poorer quality
and lower resolution
(Fig. \ref{fig:FR_datasets}), 
thus hindering the FR performance.
{\bf (2)} Deep models are highly domain-specific and likely
yield big performance degradation in cross-domain deployments.
This is particularly so when the domain gap between the training and test data 
is large, such as web and surveillance faces.
In such cases, transfer learning 
is challenging \citep{pan2010survey}.
The challenge can be further increased by the
scarcity of labelled surveillance data.
{\bf (3)} Instead of the closed-set search considered in most existing methods, 
FR in surveillance scenarios is intrinsically open-set
where the probe face ID 
is not necessarily present in the gallery. 
This brings about a significant challenge by additionally
requiring the FR system to accurately reject non-target people (i.e. distractors) 
whilst simultaneously not missing target IDs. 
Therefore, open-set FR is more challenging 
since the distractors can be 
with arbitrary variety. 

\vspace{0.1cm}
\noindent {\bf (III) Low-Resolution Face Recognition.}
A unique FR challenge in surveillance FR is low-resolution \citep{wang2014low}.
%
For FR on image pairs with large resolution difference,
the resolution discrepancy problem emerges.
Generally, existing low-resolution FR methods are fallen into two categories:
{\bf (1)} image super-resolution \citep{gunturk2003eigenface,jia2005multi,liu2005hallucinating,li2008hallucinating,hennings2008simultaneous,zou2012very,fookes2012evaluation,bilgazyev2011sparse,xu2013face,wang2016studying,jiang2016facial}, and
{\bf (2)} resolution-invariant learning \citep{ahonen2008recognition,wang2008scale,ren2012coupled,he2005neighborhood,shekhar2011synthesis,zhou2011low,abiantun2006low,lei2011local,biswas2010multidimensional,biswas2012multidimensional,li2009coupled,choi2009color,li2010low,choi2008feature}.

The first category is based on two learning criteria:
pixel-level visual fidelity and face ID discrimination.
Existing methods often conduct two training processes 
with more focus on appearance enhancement 
\citep{gunturk2003eigenface,wang2003face}.
There are recent attempts \citep{hennings2008simultaneous,zou2012very,wang2016studying} that unite
the two sub-tasks for more discriminative learning. 

The second category of methods aims to 
extract resolution-invariant features \citep{choi2009color,lei2011local,ahonen2008recognition,wang2008scale}
or learning a cross-resolution structure transformation \citep{choi2008feature,wong2010dynamic,ren2012coupled,he2005neighborhood,shekhar2011synthesis}.
As deep FR models are data driven, 
they can be conceptually categorised into this strategy
whenever suitable training data is available to model optimisation.

However, all the existing methods have a number of limitations:
(1) Considering small scale and/or down-sampled artificial low-resolution face images 
in the closed-set setting, therefore unable to reflect the genuine
surveillance FR challenge at scales.
(2) Relying on hand-crafted features and linear/shallow model structures, 
therefore subject to suboptimal generalisation.
%
(3) Requiring coupled low- and high-resolution image pairs 
from the same domain for model training, 
but unavailable in the surveillance scenarios.
%
%

\changed{In low-resolution (LR) imagery face recognition deployments,
	two typical operational settings exist.
One common setting is LR-to-HR (high-resolution) which
aims to match LR probe images against
HR gallery images such as passport or other document photos
\citep{bilgazyev2011sparse,biswas2010multidimensional,shekhar2011synthesis,biswas2012multidimensional,ren2012coupled}.
This
is a widely used approach by Law Enforcement
Agencies to matching potential candidates against a watch-list.
On the other hand, there is another operational setting
which requires LR-to-LR imagery face matching when both the
probe and the gallery images are LR facial images 
\citep{wang2003face,gunturk2003eigenface,fookes2012evaluation,zou2012very,wang2016studying}.}

\changed{
Generally, LR-to-LR face recognition occurs in a less
stringent deployment setting at larger scales when
pre-recorded (in a controlled environment) HR facial images of
a watch-list do not exist, nor there is a pre-defined watch-list.
In an urban environment, 
there is no guarantee of controlled access points to record
HR facial images of an average person in unconstrained public spaces
due to the commonly used wide field of view of
CCTV surveillance cameras and long distances between the
cameras and people. 
In large, public space video surveillance data contain a very
large number of ``joe public'' without HR facial images pre-enrolled
for face recognition. 
Video forensic analysis often requires large scale
people searching and tracking over distributed and disjoint
large spaces by face recognition of {\em a priori} unknown
(not enrolled) persons triggered by a public disturbance, when
the only available facial images are from LR CCTV videos.
More recently, a rapid emergence of smart shopping, such as
the Amazon Go, Alibaba Hema and JD 7Fresh supermarkets,
also suggests that any face recognition techniques
for individualised customer in-store localisation
(track-and-tag) cannot assume unrealistically stringent HR
facial imagery enrollment of every single potential customer
if the camera system is to be cost effective.
}

\begin{table*} 
	\centering
	\setlength{\extrarowheight}{0.5mm}
	\setlength{\tabcolsep}{0.25cm} 
	\caption{
		Person re-identification datasets 
		utilised in constructing the {\em QMUL-SurvFace}
		challenge.
	}
	\vskip -0.2cm
	\label{table:reid_dataset}
	\begin{tabular}{l||c|c|c|c|c}
		\hline
		Person Re-Identification Dataset & IDs & Detected IDs & Bodies & Detected {Faces} & Nation 
		\\
		\hline
		Shinpuhkan~\citep{kawanishi2014shinpuhkan2014} 
		& 24 & 24 & 22,504 & 6,883 & Japan 
		\\
		WARD~\citep{martinel2012re} 
		& 30 & 11 & 1,436 & 390 & Italy 
		\\
		RAiD~\citep{das2014consistent} 
		& 43 & 43 & 6,920 & 3,724 & US 
		\\
		CAVIAR4ReID~\citep{cheng2011custom} 
		& 50 & 43 & 1,221 & 141 & Portugal 
		\\
		SARC3D~\citep{baltieri2011sarc3d} 
		& 50 & 49 & 200 & 107 & Italy 
		\\
		ETHZ~\citep{schwartz09d} 
		& 148 & 110 & 8,580 & 2,681 & Switzerland 
		\\
		3DPeS~\citep{baltieri20113dpes} 
		& 192 & 133 & 1,012 & 366 & Italy 
		\\
		QMUL-GRID~\citep{loy2009multi} 
		& 250 & 242 & 1,275 & 287 & UK 
		\\
		iLIDS-VID~\citep{wang2014person} 
		& 300 & 280 & 43,800 & 14,181 & UK 
		\\
		SDU-VID~\citep{liu2015spatio} 
		& 300 & 300 & 79,058 & 67,988 & China
		\\
		PRID 450S~\citep{roth14a} 
		& 450 & 34 & 900 & 34 & Austria 
		\\
		VIPeR~\citep{gray2008viewpoint} 
		& 632 & 456 & 1,264 & 532 & US 
		\\
		CUHK03~\citep{li2014deepreid} 
		& 1,467 & 1,380 & 28,192 & 7,911 & China 
		\\
		Market-1501~\citep{zheng2015scalable} 
		& 1,501 & 1,429 & 25,261 & 9,734 & China 
		\\
		Duke4ReID \citep{gou2017dukemtmc4reid} 
		& 1,852 & 1,690 & 46,261 & 17,575 & US 
		\\
		CUHK-SYSU~\citep{xiao2016end} 
		& 8,351 & 6,694 & 22,724 & 12,526 & China 
		\\
		LPW~\citep{song2017region}
		& 4,584 & 2,655 & 590,547 & 318,447 & China 
		\\	
		\hline
		\bf Total 
		& 20,224 & 15,573 & 881,065  & {463,507} & Multiple 
		\\
		\hline
	\end{tabular}
\end{table*}

\vspace{0.1cm}
\noindent \textbf{Image Super-Resolution.}
Super-resolution methods have been significantly developed 
thanks to the strong capacity of deep CNN models 
in regressing the pixel-wise loss between reconstructed and ground-truth images
\citep{yang2014single,dong2014learning,dong2016accelerating,ledig2016photo,yu2016ultra,kim2016accurate,kim2016deeply,LapSRN,DRRN17}.
A performance summary of six state-of-the-art deep models 
on five benchmarks is given in Table~\ref{table:srperformance}.
Mostly, FR and super-resolution researches advance independently, with both assuming  
the availability of large high-resolution training data. 
In surveillance, high-resolution images are typically unavailable,
which in turn resorts the existing methods to transfer learning.
When the training and test data distributions are very different, 
super-resolution becomes extremely challenging
due to an extra need for domain adaptation.

As an object-specific super-resolution, 
face hallucination is dedicated to fidelity restoration of facial appearance 
\citep{baker2000hallucinating,wang2005hallucinating,chakrabarti2007super,liu2007face,jia2008generalized,jin2015robust,zhu2016deep,cao2017attention,yu2017hallucinating}.
A common hallucination approach is transferring high-frequency details and structure information 
from exemplar high-resolution images.
This is typically achieved by mapping low- and high-resolution training pairs.
Existing methods require noise-free input images,
whilst assuming
stringent part detection and dense correspondence alignment.
Otherwise, overwhelming artifacts may be introduced. 
These assumptions significantly limit their usability
to low-resolution surveillance data due to the presence
of uncontrolled noise and
the absence of coupled high-resolution images.
%


\section{QMUL-SurvFace Recognition Challenge}\label{sec:QMUL_SurvFace}

\subsection{A Native Low-Res Surveillance Face Dataset}
To our best knowledge, there is no large native surveillance 
FR challenge in the public domain.
To stimulate the research on this problem, 
we construct a new large scale benchmark (challenge)
by automatically extracting faces of the uncooperative general public 
appearing in real-world surveillance videos and images. 
We call this challenge \textbf{\em QMUL-SurvFace}.
Unlike most existing FR challenges using 
either high-quality web
or {simulated} 
surveillance images captured
in controlled conditions
therefore {\em failing} to evaluate the true surveillance 
FR performance,
we explore real-world native surveillance imagery data from a combination of 17 person 
re-identification benchmarks
which were collected in 
different surveillance scenarios across diverse
sites and multiple countries (Table \ref{table:reid_dataset}). 

\begin{figure} [!hbpt]
	\centering
	\begin{subfigure}[t]{0.5\textwidth}
		\includegraphics[width=.85\textwidth]{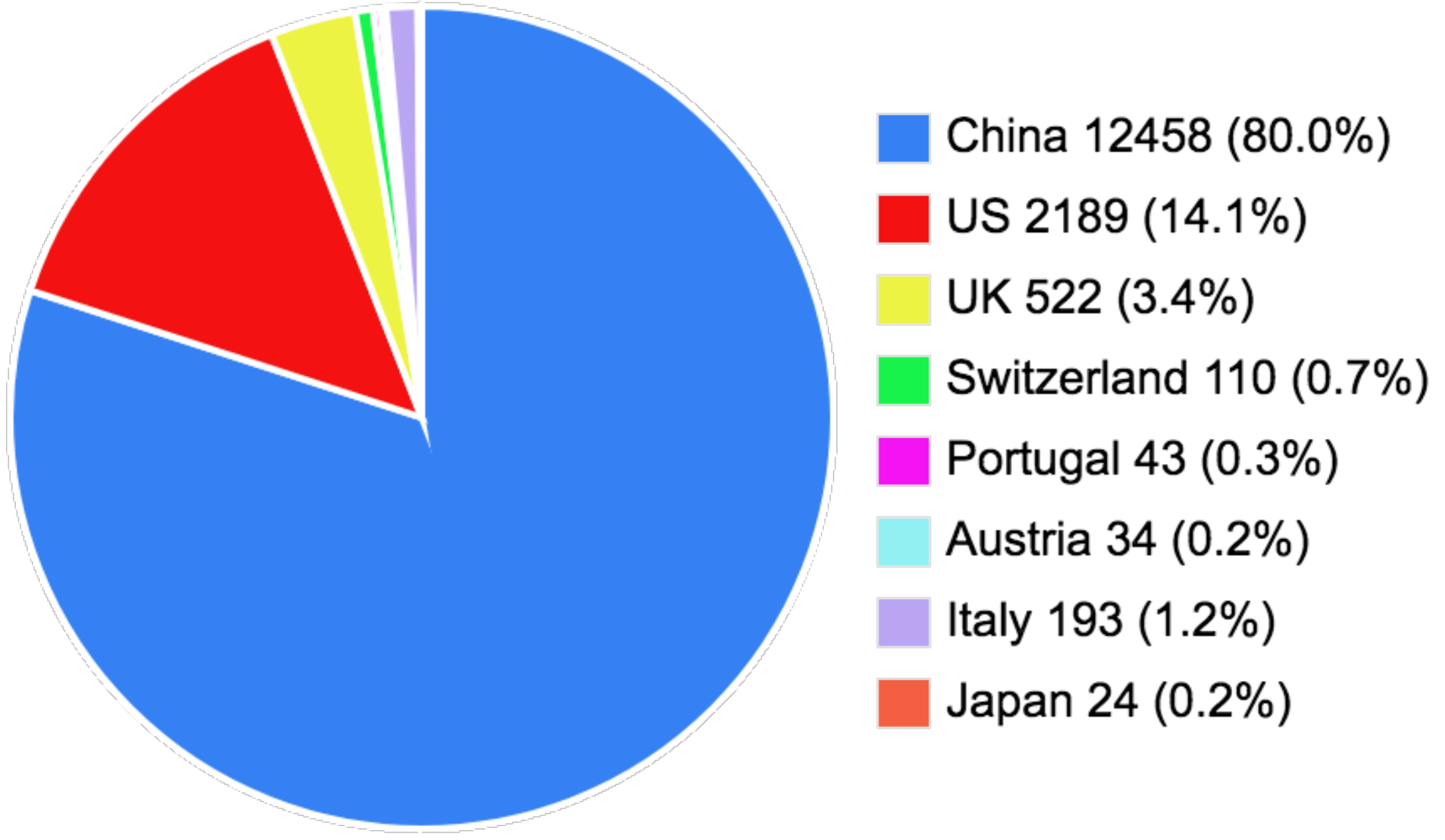}
		\vskip -0.2cm
		\caption{Geographic distribution of face IDs.}
	\end{subfigure}
	\vskip 0.5cm
	\begin{subfigure}[t]{0.5\textwidth}
		\includegraphics[width=.85\textwidth]{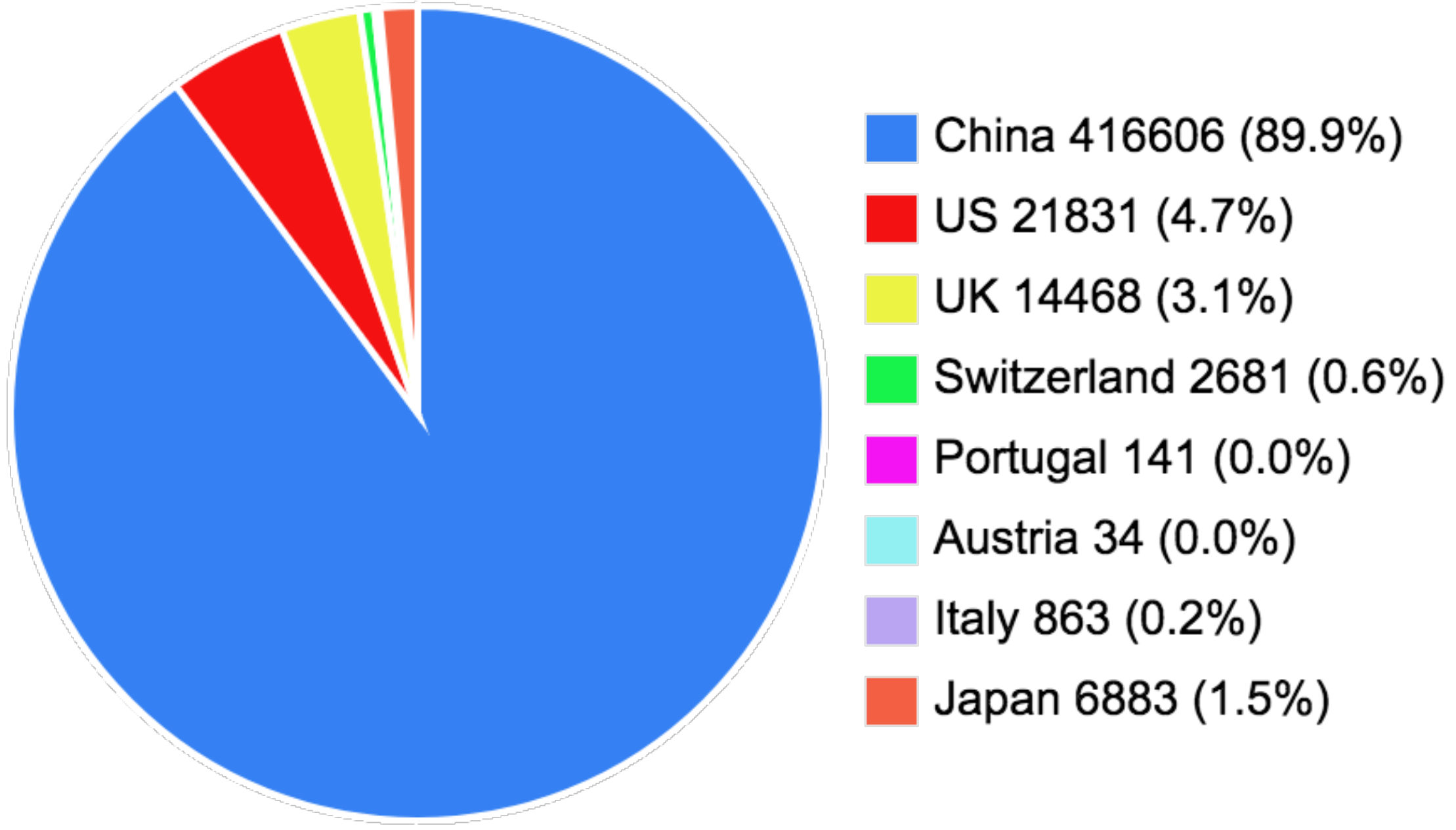}
		\vskip -0.2cm
		\caption{Geographic distribution of face images.}
	\end{subfigure}
	\caption{
		Geographic distributions of 
		(a) face IDs and (b) face images
		in the {\em QMUL-SurvFace}
		challenge.
	}
	\label{fig:geographic}
\end{figure}

\begin{figure*} [!hbpt]
	\centering
	\includegraphics[width=1\textwidth]{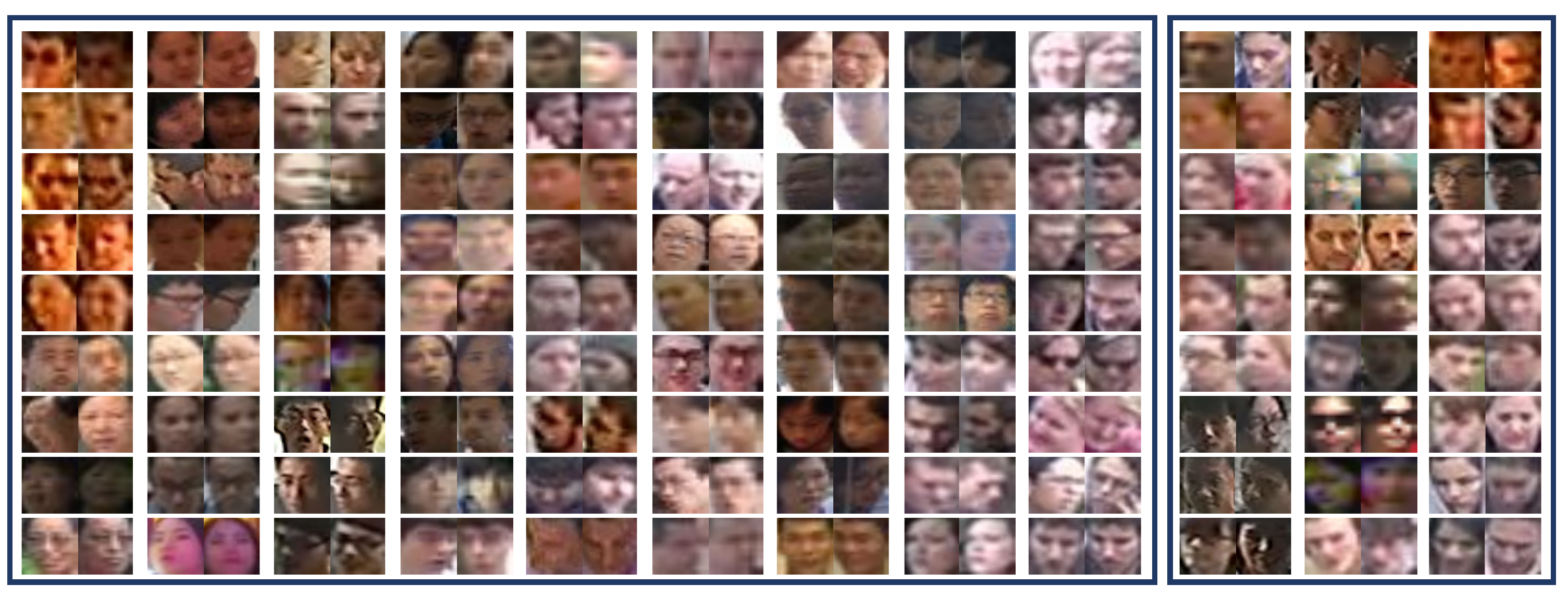}
	\vskip -0.2cm
	\caption{Example face image pairs from {\em QMUL-SurvFace}.
		{\bf Left}: matched pairs.
		{\bf Right}: unmatched pairs.}
	\label{fig:images_survface}
\end{figure*}

\vspace{0.1cm}
\noindent {\bf Dataset Statistics.}
The {\em QMUL-SurvFace} challenge contains 463,507 low-resolution 
face images of 15,573 unique person IDs
with uncontrolled appearance variations in 
pose, illumination, motion blur, occlusion and background clutter
(Fig. \ref{fig:images_survface}).
Among all, there are 10,638 (68.3\%) people 
each associated with two or more detected face images.
This is the largest native surveillance face 
benchmark to date 
(Table \ref{table:existDatasets}).

\begin{figure*} [!hbpt]
	\centering
	\includegraphics[width=1\linewidth]{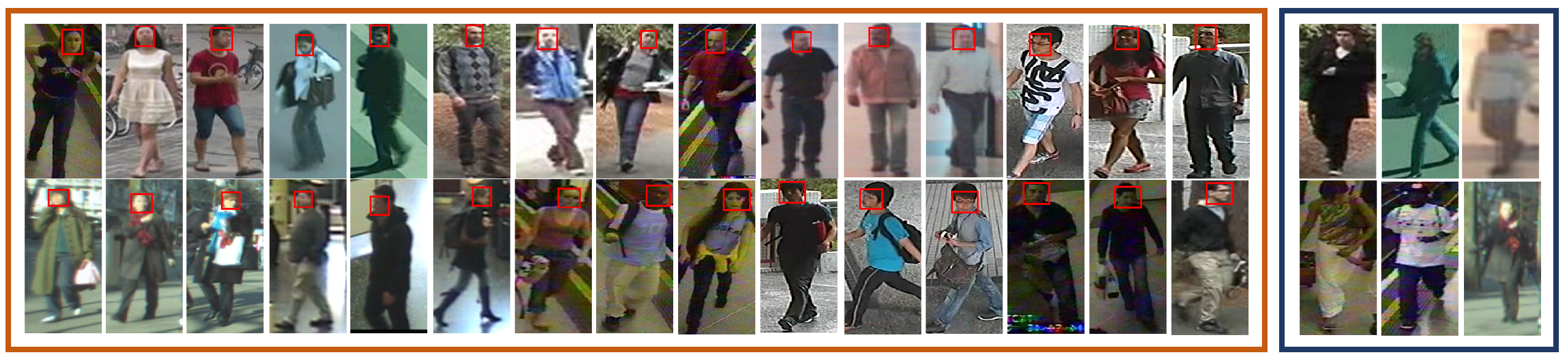}
	\vskip -0.2cm
	\caption{Illustration of face detections in native surveillance person images.
		{\bf Left}: Auto-detected faces. 
		{\bf Right}: Failure cases, the detector fails to identify the
		face due to low-resolution, motion blur, extreme pose, poor illumination, and background clutters.}
	\label{fig:det_survface}
\end{figure*}

\vspace{0.1cm}
\noindent {\bf Face Image Collection.}
To enable QMUL-SurvFace represent a scalable test scenario, 
we automatically extracted the facial images by deploying 
the TinyFace detector~\citep{hu2016finding}
on re-identification surveillance data (Fig. \ref{fig:det_survface}).
Manually labelling faces is
non-scalable due to the huge amount of surveillance video data.
Note that not all faces in the source images can be successfully detected 
given imperfect detection, 
poor image quality, and extreme head poses.
The average face detection recall is 77.0\% 
(15,573 out of 20,224)
in ID 
and 
52.6\% (463,507 out of 881,065)
in image.
%
%
Face detection statistics across all person re-identification datasets 
are summarised
in Table \ref{table:reid_dataset}.

\vspace{0.1cm}
\noindent {\bf Face Image Cleaning and Annotation.}
To make an accurate FR challenge,
we manually cleaned QMUL-SurvFace data
by filtering out false detections.
We manually thrown away all movie and TV program (non-surveillance) 
image data in the CUHK-SYSU dataset
%
These were labelled by two independent annotators and 
a subsequent mutual cross-check.
For face ID annotation, we used the 
person labels available in sources 
where we assume no ID overlap across datasets.
This is rational since they were independently created over different
time and surveillance venues, that is, the possibility that a person appearing 
in multiple re-identification datasets is extremely low.

\begin{figure}[!hbpt]
	\centering
	\includegraphics[width=.47\textwidth]{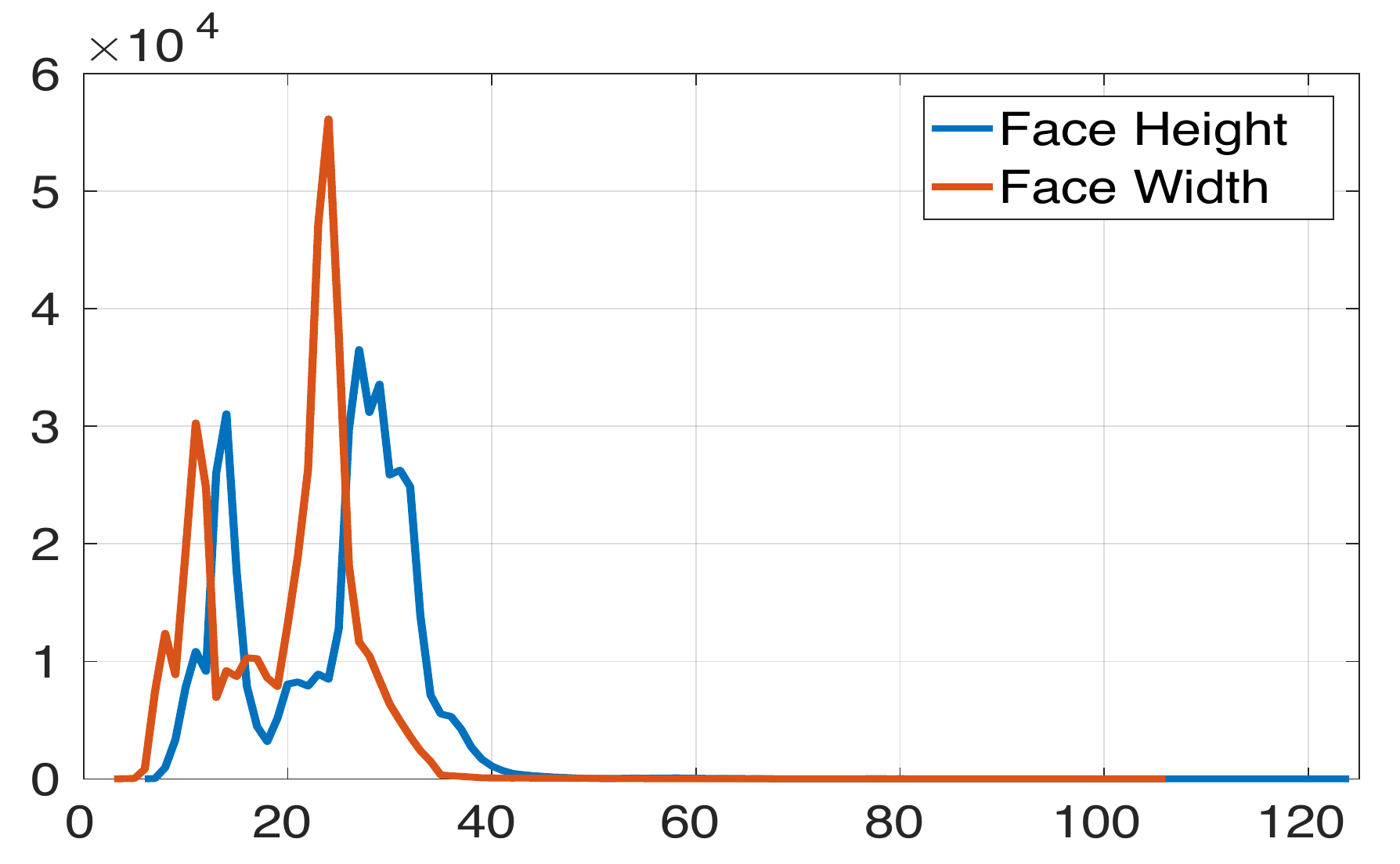}
	\vskip -0.2cm
	\caption{
		Scale distributions of {\em QMUL-SurvFace} images.
	}
	\label{fig:image_size_dis}
\end{figure}

\begin{figure}[!hbpt]
	\centering
	\includegraphics[width=1\linewidth]{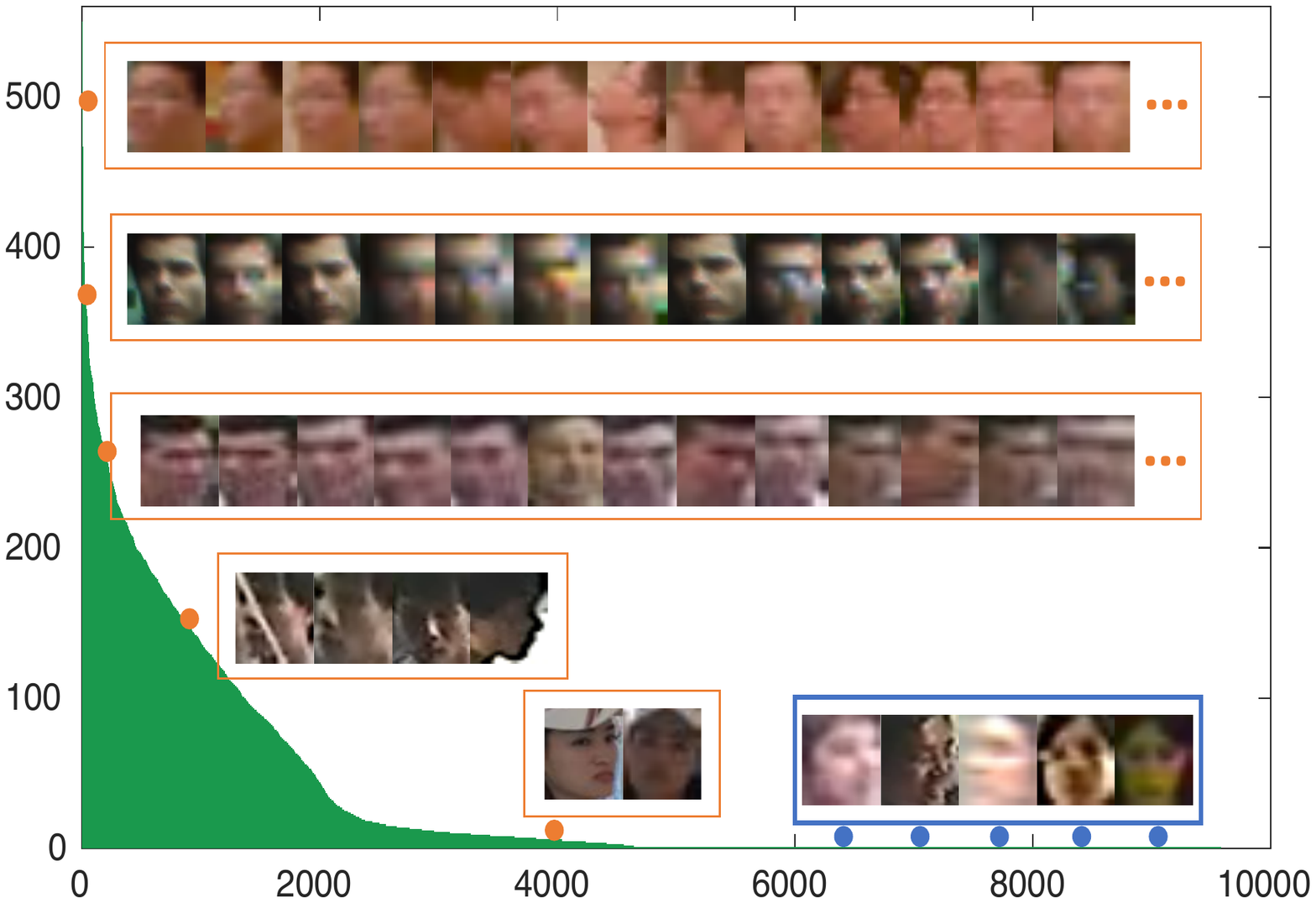}
	\vskip -0.2cm
	\caption{
		Image frequency over all {\em QMUL-SurvFace} IDs. 
	}
	\label{fig:image_num_dis}
\end{figure}

\vspace{0.1cm}
\noindent {\bf Face Characteristics.}
In contrast to existing FR challenges,
QMUL-SurvFace is uniquely characterised by
very low resolution faces typical in video surveillance
(Fig. \ref{fig:image_size_dis})
-- one major source that makes surveillance FR challenging
\citep{wang2014low}.
The face spatial resolution ranges from 6/5 to 124/106 pixels
in height/width, and the average is 24/20.
%
This dataset exhibits a power-law distribution in frequency
ranging from 1 to 558 (Fig. \ref{fig:image_num_dis}).
Besides, the QMUL-SurvFace people 
feature a wide variation in geographic origin\footnote{
In computing the nationality statistics, 
it is assumed that all people of a person re-identification 
dataset share the same nationality.
Per-ID nationality labels are not available.}
(Fig. \ref{fig:geographic}).
Given low-resolution facial appearance,
surveillance FR may be less sensitive 
to the nationality than 
high-resolution web FR.

\begin{figure*} [!hbpt]
	\centering
	\includegraphics[width=1\textwidth]{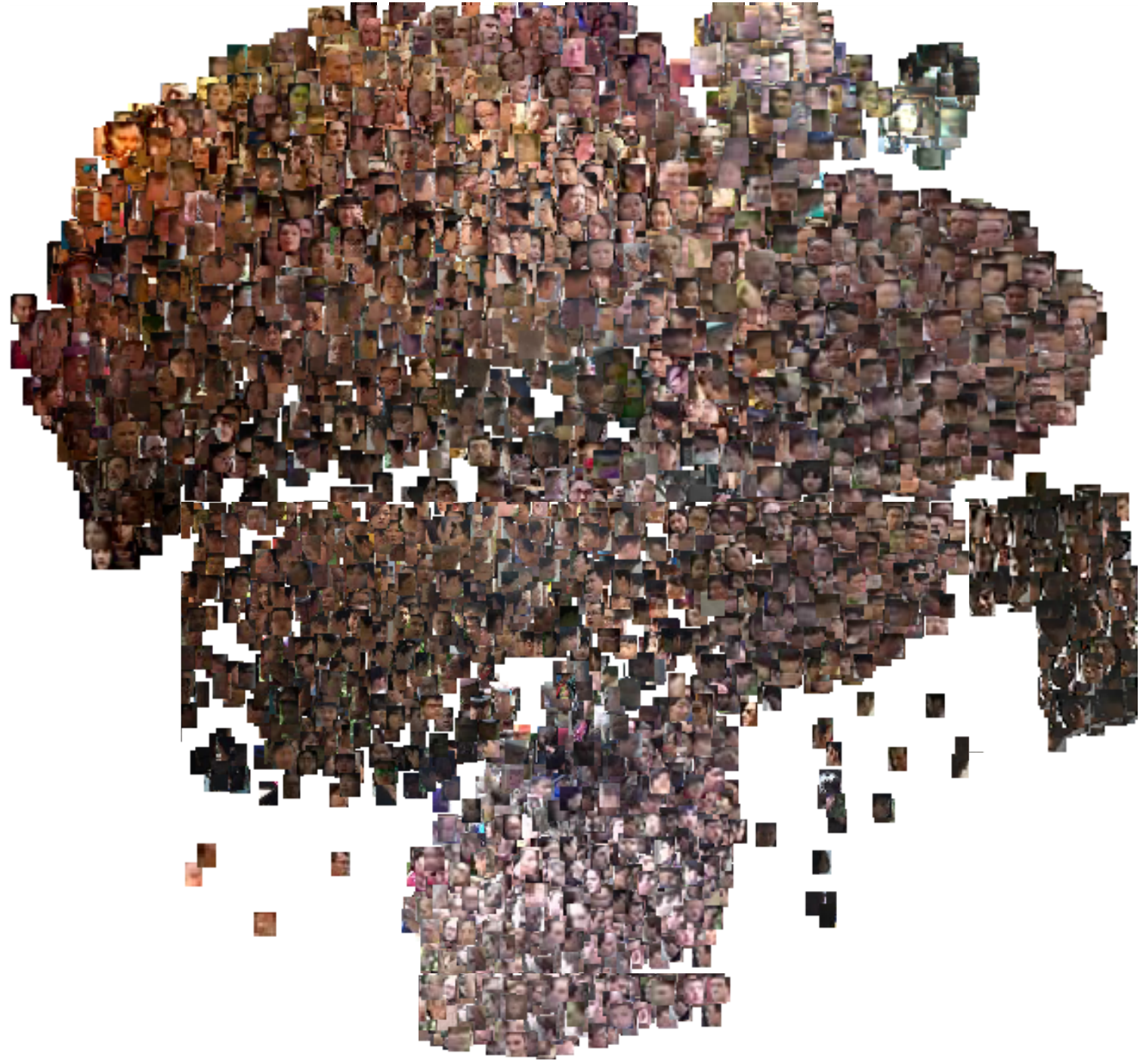}
	\vskip -0.2cm
	\caption{A glimpse of native surveillance face images from the {\em QMUL-SurvFace} challenge. }
	\label{fig:images_survface_group}
\end{figure*}

\begin{table} 
	\centering
	\setlength{\extrarowheight}{0.5mm}
	\setlength{\tabcolsep}{0.05cm} 
	\caption{
		Benchmark data partition of {\em QMUL-SurvFace}.
		Numbers in parentheses: per-identity image size range.
	}
	\vskip -0.2cm
	\label{table:survface_datasplit}
	\begin{tabular}{c||c|c|c}
		\cline{1-4}
		{Split} & All & Training & Test
		\\
		\cline{1-4}
		IDs  & 15,573 & 5,319 & 10,254
		\\
		\cline{1-4}
		Images & 463,507 (1$\sim$558) & 220,890 (2$\sim$558) & 242,617 (1$\sim$482)
		\\
		\cline{1-4}
		\hline
	\end{tabular}
\end{table}


\subsection{Evaluation Protocols}


FR performance are typically evaluated with two applications
(verification and identification)
and two protocols (closed-set and open-set).

\vspace{0.1cm}
\noindent \textbf{Data Partition.}
To benchmark the evaluation protocol, we first need to split 
the {\em QMUL-SurvFace} data into training and test sets.
We divide the 10,638 IDs each with $\geq 2$ face images into two halves: 
one half (5,319) for training,
one half (5,319) plus the remaining 4,935 single-shot IDs (in total 10,254) for test
(Table \ref{table:survface_datasplit}).
We benchmark only {\em one} train/test data split 
since the dataset is sufficiently large in ID classes and face images to support 
a statistically stable evaluation.

We apply the same data partition as above for both face identification and verification.
%
All face images of training IDs are used to train FR models. 
%
Additional imagery from other sources 
may be used 
subject to no facial images of test IDs.
%
The use of test data depends on 
the application and protocol. 
%
Next, we present the approach for benchmarking {\em face identification}
and {\em face verification} on {QMUL-SurvFace}.

\begin{table} 
	\centering
	\setlength{\extrarowheight}{0.5mm}
	\setlength{\tabcolsep}{0.2cm} 
	\caption{
		Benchmark face verification and identification protocols on {\em QMUL-SurvFace}.
		TAR: {True Accept Rate};
		FAR: {False Accept Rate};
		ROC: Receiver Operating Characteristic;
		FPIR: {False Positive Identification Rate};
		TPIR: {True Positive Identification Rate}.
	}
	\vskip -0.2cm
	\label{table:survface_veri}
	\begin{tabular}{c||c|c||c}
		\hline
		\multicolumn{4}{c}{\em \bf 1:1 Face Verification Protocol} \\ 
		\hline
		Matched Pairs & 5,319  & Unmatched Pairs & 5,319
		\\ \hline
		Metrics & \multicolumn{3}{c}{TAR@FAR, ROC} \\
		\hline 
		\hline
		\multicolumn{4}{c}{\em  \bf 1:N Face Identification Protocol} \\
		\hline
		Scenario &  \multicolumn{3}{c}{Open-Set} \\
		\hline
		Partition & Probe & \multicolumn{2}{c}{Gallery} \\
		\hline
		IDs &  10,254 & \multicolumn{2}{c}{3,000} \\ \hline
		Images &  182,323 & \multicolumn{2}{c}{60,294} \\ \hline
		Metrics & \multicolumn{3}{c}{TPIR@FPIR, ROC} \\
		\hline
	\end{tabular}
\end{table}

\vspace{0.1cm}
\noindent {\bf Face Verification.}
The verification protocol measures the comparing performance 
of face pairs.
Most FR methods evaluated on LFW adopt 
this protocol.
Specifically, one presents a face image to a FR system with a claimed ID
represented by an enrolled face. 
The system accepts the claim if their matching similarity score
is greater than a threshold $t$,
or rejects otherwise \citep{phillips2010frvt}.
%
The protocol specifies the sets of {\em matched} and {\em unmatched} 
pairs that FR methods should perform in evaluation.
For each test ID, we generate one matched pair, 
i.e. a total of 5,319 pairs (Table \ref{table:survface_veri}).
We generate the same number (5,319) of unmatched pairs
by randomly sampling between a face and nonmated ones.
For performance measurement, each of these pairs is to be evaluated
by computing a matching similarity score.

In the verification process, two types of error can occur: 
(1) {A false accept} -- a distractor claims an ID of interest; 
(2) {A false reject} -- the system mistakenly declines the ID of interest. 
As such, we define the False Accept Rate (FAR) as 
the fraction of unmatched pairs with the corresponding score $s$ above threshold $t$
\begin{equation}
\text{FAR}(t) = \frac{|\{s \geq t, \; \text{where} \; s \in U \}|}{|U|}
\label{eqn:FAR}
\end{equation}
where $U$ denotes the set of unmatched pairs.
In contrast, the False Rejection Rate (FRR) represents 
the fraction of
matched pairs with matching score $s$ below a threshold $t$:
\begin{equation}
\text{FRR}(t) = \frac{|\{s < t, \; \text{where} \; s \in M\}|}{|M|}
\label{eqn:FRR}
\end{equation}
where $M$ is the set of matched pairs. For understanding convenience,
we further define the True Accept Rate (TAR), the complement of FRR, as
\begin{equation}
\text{TAR}(t) = 1 - \text{FRR}(t).
\label{eqn:TAR}
\end{equation}
For face verification evaluation on QMUL-SurvFace, we use 
the paired TAR@FAR measure.

We also utilise the receiver operating characteristic (ROC)
analysis measurement
by varying the threshold $t$ and generating 
a TAR-vs-FAR ROC curve.
The overall accuracy performance can be measured by
the area under the ROC curve, which is abbreviated as AUC.
See the top half of Table \ref{table:survface_veri} for 
the verification protocol summary.

\vspace{0.1cm}
\noindent {\bf Face Identification.}
In forensic and surveillance applications however, 
it is face identification that is of more interest
\citep{best2014unconstrained,ortiz2014face},
%
and arguably a more intricate and non-trivial problem
since a probe image must be compared against all gallery IDs 
\citep{wang2016face,kemelmacher2016megaface}.

Most existing FR methods in the literature consider the {\em closed-set} scenario,
assuming that each probe subject is present in the gallery.
We construct the evaluation setup for the closed-set scenario 
on QMUL-SurvFace 
through the following process.
For each of the 5,319 multi-shot test IDs, 
we randomly sample the corresponding images into probe or gallery. 
The gallery set represents imagery involved in an operational database,
e.g. access control system's repository.
For any unique person, 
we generate a single ID-specific face template from 
one or multiple gallery images \citep{klare2015pushing}.
This makes the ranking list concise and more efficient 
for post-rank manual validation,
e.g. no case that a single ID takes multiple ranks.
The probe set represents imagery used 
to query a face identification system.

For performance evaluation in {\em closed-set} identification, 
we select the
{Cumulative Matching Characteristic} (CMC) \citep{klare2015pushing}
measure.
CMC reports the fraction of searches returning the mate (true match) 
at rank $r$ or better, with the rank-1 rate as the most common summary indicator of an algorithm's efficacy.
It is a non-threshold rank based metric.
Formally, the CMC at rank $r$ is defined as:
\begin{equation}
\text{CMC}(r) = \sum_{i=1}^{r} \frac{N_\text{mate}(i)}{N}
\label{eq:CMC}
\end{equation} 
where $N_\text{mate}(i)$ denotes the number of probe images 
with the mate ranked at position $i$, and $N$ the total probe number.

In realistic surveillance applications, however, 
most faces captured by CCTV cameras are not of any gallery person 
therefore should be detected as unknown, 
leading to the {\em open-set} protocol \citep{grother2014face,liao2014benchmark}.
This is often referred to the {\em watch-list identification} (forensic search) scenario
where only persons of interest are enrolled into the gallery,
typically each ID with several different images such as the FBI’s most wanted list\footnote{\url{www.fbi.gov/wanted}}.
To allow for the {\em open-set} surveillance FR test, 
we construct a watch list identification protocol where
only face IDs of interest are enrolled in the gallery.
specifically, we create the following probe and gallery sets:
(1) Out of the 5,319 multi-shot test IDs, 
we randomly select 3,000 and sample half face images for each selected ID
into the gallery set, i.e. the watch list.
(2) All the remaining images including these single-shot ID imagery
are used to form the probe set.
As such, the majority of probe people are {\em unknown} (not enrolled gallery IDs),
more accurately reflecting the open space forensic search nature. 

For the {\em open-set} FR performance evaluation, 
we must quantify two error types \citep{grother2014face}.
The first type is
{\em false alarm} -- 
a face image from an unknown person (i.e. nonmate search) 
is incorrectly associated with one or more enrollees' data.
This error is quantified by
the {\em False Positive Identification Rate} (FPIR):
\begin{equation}
\text{FPIR}(t) = \frac{N_\text{nm}^\text{m}}{N_\text{nm}}
\label{eq:FPIR}
\end{equation}
which measures 
the proportion of nonmate searches $N_\text{nm}^\text{m}$ (i.e. no mate faces in the gallery) that produce 
one or more enrolled candidates at or above a threshold $t$ (i.e. false alarm),
among a total of $N_\text{nm}$ nonmate searches attempted.
The second type of error is {\em miss} --
a search of an enrolled target person’s data (i.e. mate search) does not return the correct ID.
We quantify the miss error by the {\em False Negative Identification Rate} (FNIR):
\begin{equation}
\text{FNIR}(t,r) = \frac{N_\text{m}^\text{nm}}{N_\text{m}}
\label{eq:FNIR}
\end{equation}
which is the proportion of mate searches $N_\text{m}^\text{nm}$ 
(i.e. with mate faces present in the gallery) with enrolled mate found outside top $r$ ranks 
or matching similarity score below the threshold $t$,
among $N_\text{m}$ mate searches.
By default, we set $r\!=\!20$ (i.e. $\text{FNIR}(t,20)$)
which assumes a small workload by a human reviewer
employed to review the candidates returned from an identification search
\citep{grother2014face}.
In practice, a more intuitive measure may be the ``hit rate'' or {\em True Positive Identification Rate} (TPIR):
\begin{equation}
\text{TPIR}(t,r) = 1 - \text{FNIR}(t,r) 
\label{eq:TPIR}
\end{equation}
which is the complement of FNIR offering a positive statement of how often mated searches are succeeded.
In QMUL-SurvFace, we adopt the TPIR@FPIR measure as the open-set face identification performance metrics.
TPIR-vs-FPIR can similarly generate an ROC curve,
the AUC of which stands for an overall measurement
(see the bottom half of Table \ref{table:survface_veri}).

\vspace{0.1cm}
\noindent \textbf{\em Link of open-set and closed-set.}
The aforementioned performance metrics of closed-set and open-set FR 
are not completely independent but correlated. 
The $\text{CMC}(r)$ (Eqn. \eqref{eq:CMC}) can be regarded as
a special case of $\text{TPIR}(t,r)$ (Eqn. \eqref{eq:TPIR}) 
with ignored similarity scores by relaxing the threshold requirement $t$ as:
\begin{equation}
\text{CMC}(r) = \text{TPIR}(t,r)
\label{eq:CMC-TPIR}
\end{equation}
This metrics linkage is useful in enabling performance comparisons
between closed-set and open-set. 

\vspace{0.1cm}
\noindent \textbf{Considerations.}
In the literature, existing FR challenges 
adopt the closed-set evaluation protocol 
including MegaFace \citep{kemelmacher2016megaface}.
While being able to evaluate FR model generalisation capability 
in large scale search 
(e.g. 1 million gallery images in MegaFace), 
it does not fully confirm to the surveillance FR operation.
%
For surveillance FR,
human operators are often assigned with a list of target people for 
a mission
with their face images enrolled in a working system. 
The FR task is then to search the targets in 
public spaces against the gallery face images.
This is an open-set FR problem.
As a consequence, we adopt the open-set identification protocol 
as the main setting of QMUL-SurvFace (Table \ref{table:survface_veri}).
Besides, we still consider closed-set FR
experiments to enable like-for-like comparisons with existing benchmarks.

\section{Benchmark Evaluation}\label{sec:setup}
In this section, we describe two categories of techniques used in
the benchmark evaluations: 
deep learning FR and image super-resolution methods.

\subsection{Face Recognition Models}

We present the formulation details of five representative FR models 
(Table \ref{table:LFW}): 
DeepID2~\citep{sun2014deepid2}, 
CentreFace~\citep{wen2016discriminative}, 
FaceNet~\citep{schroff2015facenet},
VggFace~\citep{parkhi2015deep}, 
and 
SphereFace~\citep{liu2017sphereface}.
All the methods are based on CNN architecture
each with different objective loss function and network designs.
They are designed for 
learning a discriminative face representation space 
by increasing 
inter-ID variations whilst decreasing 
intra-ID variations.
Both variations are intrinsically complex and highly non-linear
due to that faces of the same ID may appear very differently
in varying conditions
whereas faces of different IDs may look alike. 
Once a deep FR model is trained by a standard 
SGD algorithm \citep{rumelhart1988learning,bottou2010large},
we deploy it as a feature extractor and perform FR
with $L_2$ distance.

\vspace{0.1cm}
The {\bf DeepID2 } model \citep{sun2014deepid2} is characterised by
simultaneously learning face identification and verification supervision. 
Identification is to classify a face image into one ID class by softmax cross-entropy loss \citep{krizhevsky2012imagenet}. Formally, 
we predict the posterior probability $\tilde{y}_i$ of a face image $\bm{I}_i$
over the ground-truth ID class $y_i$ among a total $n_\text{id}$ distinct training IDs:
\begin{equation}
{p}(\tilde{y}_{i} = y_i | \bm{I}_{i}) = \frac{\exp(\bm{w}_{y_i}^{\top} \bm{x}_{i})} {\sum_{k=1}^{|n_\text{id}|} \exp(\bm{w}_{k}^{\top} \bm{x}_{i})}
\label{eq:prob}
\end{equation}
where $\bm{x}_{i}$ specifies the DeepID2 feature vector of $\bm{I}_i$,
and $\bm{W}_k$ the prediction parameter
of the $k$-th ID class.
The identification training loss 
is defined as:
\begin{equation}
l_\text{id} = 
- \log \Big(p(\tilde{y}_{i} = y_i|\bm{I}_{i}) \Big)
\label{eq:loss_cls}
\end{equation}
%
%
The verification signal encourages the DeepID2 features extracted from 
the same-ID face images to be similar so to reduce the intra-person variations.
This is achieved by the pairwise contrastive loss \citep{hadsell2006dimensionality}:
\begin{equation}
l_\text{ve} = 
\left\{
\begin{array}{l l}
\frac{1}{2} \| \bm{x}_i - \bm{x}_j \|^2_2 & \;\;\; \text{if same ID,} \\
\frac{1}{2} \max \big(0, m-\| \bm{x}_i - \bm{x}_j \|^2_2 \big)^2	& \;\;\; \text{otherwise.} \\
\end{array} \right.
\label{eqn:veri}
\end{equation}
where $m$ represents the discriminative ID class margin.
The final DeepID2 model loss function is a weighted summation of the above two as:
\begin{equation}
\mathcal{L}_\text{DeepID2} = 
l_\text{id} + \lambda_\text{bln} l_\text{ve}
\label{eqn:deepid2}
\end{equation}
where $\lambda_\text{bln}$ represents the balancing hyper-parameter.
A customised 5-layers CNN is used in the DeepID2.

\vspace{0.1cm}
The {\bf CentreFace } model \citep{wen2016discriminative} also adopts the
softmax cross-entropy loss function (Eqn. \eqref{eq:loss_cls}) 
to learn inter-class discrimination.
However, it seeks for intra-ID compactness in a class-wise manner
by posing a representation constraint 
that all face image features be close to the corresponding ID centre 
as possible.
Learning this class compactness is accomplished by a centre loss function defined as: 
\begin{equation}
l_\text{centre}= 
\frac{1}{2} \| \bm{x}_i-\bm{c}_{y_i} \|_2^2
\label{eqn:centre}
\end{equation}
where $y_i$ denotes the ID class of face images $\bm{x}_i$
and $\bm{c}_{y_i}$ the up-to-date feature centre of the class $y_i$. 
As such, all face images of the same ID are constrained to
group together so that the intra-person variations 
can be suppressed.
The final loss function is integrated with the identification supervision as:
\begin{equation}
\mathcal{L}_\text{CentreFace} = l_\text{id} + \lambda_\text{bln} l_\text{centre}
\label{eqn:centre_final}
\end{equation}
Since the feature space is dynamic in the course of training, 
all class centres are progressively undated on-the-fly.
The CentreFace model is implemented in a 28-layers ResNet architecture \citep{he2016deep}.

\vspace{0.1cm}
The {\bf FaceNet }  model \citep{schroff2015facenet} sues a triplet loss function \citep{liu2009learning}
to learn a binary-class (positive {\em vs} negative pairs) feature embedding.
The triplet loss is to induce a discriminative margin 
between positive and negative pairs, defined as:
\begin{equation}
\begin{aligned}
l_\text{tri} = \max\left\{0, \; \alpha-\|\bm{x}_a - \bm{x}_n\|_2^2 +\| \bm{x}_a - \bm{x}_p\|_2^2\right\}, \\ 
\textit{subject to:}\; (\bm{x}_a,\bm{x}_p,\bm{x}_n) \in \mathcal{T}
\end{aligned}
\label{eqn:triplet}
\end{equation}
where $\mathcal{T}$ denotes the set of triplets generated based on ID labels, and 
$\alpha$ is a pre-fixed margin for separating positive ($\bm{x}_a$, $\bm{x}_p$) and negative
($\bm{x}_a$, $\bm{x}_n$) training pairs. 
By doing so, the face images for one training ID are constrained to populate on an isolated manifold against other IDs by a certain distance therefore 
posing a discrimination capability.
For fast convergence, it is critical to use triplets 
that violate the triplet constraint (Eqn. \eqref{eqn:triplet}).
To achieve this in a scalable manner, we select hard positives and negatives 
within a mini-batch. 
In our FaceNet implementation, 
an Inception-ResNet CNN architecture \citep{szegedy2017inception}
is used as a stronger replacement of the originally adopted ZF CNN \citep{zeiler2014visualizing}.

\vspace{0.1cm}
The {\bf VggFace} model \citep{parkhi2015deep} considers both 
identification and triplet training schemes in a sequential manner.
Specifically, we first train the model by a softmax cross-entropy loss (Eqn. \eqref{eq:loss_cls}).
We then learn the feature embedding by a triplet loss (Eqn. \eqref{eqn:triplet})
where only the last full-connected layer is updated to implement
a discriminative projection. 
A similar hard sample mining strategy is applied in the second step for more efficient optimisation.
The VggFace adopts a 16-layers VGG16 CNN \citep{simonyan2014very}.

\vspace{0.1cm}
The {\bf SphereFace} model \citep{liu2017sphereface} 
exploits a newly designed angular margin based 
softmax loss function. 
This loss differs from Euclidean distance based triplet loss (Eqn. \eqref{eqn:triplet}) 
by performing feature discrimination learning in a hyper-sphere manifold.
The motivation is that, multi-class features learned by the identification loss 
exhibit an intrinsic angular distribution.
Formally, the angular softmax loss is formulated as:
\begin{equation}
\begin{aligned}
l_\text{ang} = -\log \Big(\frac{e^{\|\bm{x}_i\|\psi(\theta_{{y_i},i})}}{e^{\|\bm{x}_i\|\psi(\theta_{{y_i},i})}+\sum_{j\neq y_i}e^{\|\bm{x}_i\| \psi(\theta_{j,i})}} \Big), \\
\text{where} \;\; \psi(\theta_{y_i},i)=(-1)^k \cos(m\theta_{y_i},i)-2k, \\ 
\text{subject to:} \;\; 
\theta_{{y_i},i}\in[\frac{k\pi}{m},\frac{(k+1)\pi}{m}], 
\;\; k \in [0, m-1]  \\
\end{aligned}
\end{equation}
where $\theta_{j,i}$ specifies the angle between normalised identification weight
$\bm{W}_j$ ($\|\bm{W}_j\| = 1$) for $j$-th class and training sample $\bm{x}_i$,
$m$ ($m\geq 2$) the pre-set angular margin,
and $y_i$ the ground-truth class of $\bm{x}_i$.
Specifically, this design manipulates the angular decision boundaries 
between classes and enforces a constraint
$\cos(m\theta_{y_i}) > \cos(\theta_j)$ for any $j \neq y_i$.
When $m\geq 2$ and $\theta_{y_i} \in [0, \frac{\pi}{m}]$,
this inequation $\cos(\theta_{y_i}) > \cos(m\theta_{y_i})$ holds.
Therefore, $\cos(m\theta_{y_i})$ represents a lower boundary of $\cos(\theta_{y_i})$
with larger $m$ leading to a wider angular inter-class margin. 
Similar to CentreFace, a 28-layers ResNet CNN is adopted in
the SphereFace implementation.

\subsection{Image Super-Resolution Models}
\label{sec:method_SR}

We present the formulation details of five image super-resolution methods 
(Table \ref{table:srperformance}):
SRCNN~\citep{dong2014learning},
FSRCNN~\citep{dong2016accelerating},
LapSRN~\citep{LapSRN},
VDSR~\citep{kim2016accurate}, and
DRRN~\citep{DRRN17}.
Similar to FR methods above,
these super-resolution models exploit CNN architectures.
%
A super-resolution model aims to
learn a highly non-linear mapping between low-
and high-resolution images. 
This requires 
ground-truth low- and high-resolution training pairs.
Once a model is trained,
we deploy it to restore poor resolution surveillance faces before 
performing FR.

\vspace{0.1cm}
The {\bf SRCNN} model \citep{dong2014learning}
is one of the first deep methods
achieving remarkable success in super-resolution.
The design is motivated by
earlier sparse-coding based methods \citep{yang2010image,kim2010single}.
By taking the end-to-end learning advantage of neural networks,
SRCNN formulates originally separated components 
in a unified framework
to realise a better mapping function learning. 
A mean squared error (MSE) is adopted as the loss function:
%
\begin{equation}
l_\text{mse}= \| f(\bm{I}^\text{lr}; \bm{\theta}) - \bm{I}^\text{hr} \|_2^2
\label{eq:MSE}
\end{equation}
where $\bm{I}^\text{lr}$ and $\bm{I}^\text{hr}$ denotes 
coupled low- and high-resolution training images,
and $f()$ the to-be-learned super-resolution function
with the parameters denoted by $\bm{\theta}$.
This model takes bicubic interpolated images
as input. 

\vspace{0.1cm}
The {\bf FSRCNN} model \citep{dong2016accelerating} is an accelerated and 
more accurate variant of SRCNN \citep{dong2014learning}.
This is achieved by taking the original low-resolution images as input,
designing a deeper hourglass (shrinking-then-expanding) shaped non-linear mapping module,
and adopting a deconvolutional layer 
for upscaling the input.
The MSR loss function (Eqn. \eqref{eq:MSE}) is used for training.

\vspace{0.1cm}
The {\bf VDSR} model~\citep{kim2016accurate} 
improves over SRCNN \citep{dong2014learning} 
by raising the network depth from $3$ to $20$ convolutional layers.
%
The rational of deeper cascaded network design
is 
to exploit richer contextual information over large image regions (e.g. 41$\times$41 in pixel) 
for enhancing high frequency detail inference.
To effectively train this model,
the residual learning scheme is adopted.
That is, the model is optimised to learn 
a residual image between the input and 
ground-truth high-resolution images.
VDSR is supervised by
the MSE loss (Eqn. \eqref{eq:MSE}).


\vspace{0.1cm}
The {\bf DRRN} model~\citep{DRRN17}
constructs an even deeper (52-layers) network
by jointly exploiting residual and recursive learning.
%
In particular, except for the global residual learning between the input and output as VDSR,
this method also exploits local residual learning via short-distance ID branches
to mitigate the information loss across all the layers.
This leads to a multi-path structured network module.
Inspired by \citep{DRRN17}, all modules share the parameters and input so that multiple recursions
can be performed 
in an iterative fashion without increasing the model parameter size.
The MSE loss function (Eqn. \eqref{eq:MSE}) is used to 
supervise in model training. 

\vspace{0.1cm}
The {\bf LapSRN} model~\citep{LapSRN} consists in
a multi-levels of cascaded sub-networks 
designed to progressively predict high-resolution 
reconstructions in a coarse-to-fine fashion.
This scheme is hence contrary to the four one-step reconstruction models above.
Same as VDSR and DRRN, the residual learning scheme is exploited
along with an upscaling mapping function to
alleviate the training difficulty whilst enjoying
more discriminative learning. 
For training, it adopts a Charbonnier penalty function~\citep{bruhn2005lucas}:
\begin{equation}
\begin{aligned}
l_\text{cpf} =  \sqrt{\|f(\bm{I}^\text{lr}; \bm{\theta}) - \bm{I}^\text{hr} \|_2^2+\varepsilon^2}
\end{aligned}
\label{eqn:CPF}
\end{equation}
where $\varepsilon$ (e.g. set to $10^{-3}$) is a pre-fixed noise constant.
Compared to MST, this loss 
has a better potential to suppress training outliers. 
Each level is supervised concurrently with a separate loss against the corresponding ground-truth 
high-resolution images. 
This multi-loss structure resembles the benefits of deeply-supervised models
\citep{lee2015deeply,xie2015holistically}.

%
{\em Discussion.} It is worth pointing out that using the MSE loss to train a super-resolution model 
favours the Peak Single-to-Noise Ratio (PSNR) rate, a common
performance measurement.
This suits deep neural networks as MSE is differentiable, 
but cannot guarantee the perceptual quality.
One phenomenon is that 
super-resolved images by an MSE supervised model
are likely to be overly-smoothed and blurred.
We will evaluate the super-resolution models
(Sec. \ref{sec:eval_SR_FR}).

\begin{table*} [h] 
	\centering
	\setlength{\tabcolsep}{0.13cm} 
	\caption{Face identification results on {\em QMUL-SurvFace}.
		{Protocol}: Open-Set. 
		{Metrics}: TPIR$20$@FPIR ($r\!=\!20$) and AUC.
		``-'': No results available due to failure of model convergence.
	}
	\vskip -0.2cm
	\label{table:results_iden_open_survface}
	\begin{tabular}{c||cccc|c|cccc|c|cccc|c}
		\hline 
		Training Data & \multicolumn{5}{c|}{QMUL-SurvFace} 
		& \multicolumn{5}{c|}{CASIA \citep{liu2015faceattributes}}
		& \multicolumn{5}{c}{CASIA + QMUL-SurvFace} \\
		\hline
		\multirow{2}{*}{Metrics}
		& \multicolumn{4}{c|}{TPIR$20$(\%)@FPIR} & \multirow{2}{*}{AUC (\%)}
		& \multicolumn{4}{c|}{TPIR$20$(\%)@FPIR} & \multirow{2}{*}{AUC (\%)}
		& \multicolumn{4}{c|}{TPIR$20$(\%)@FPIR} & \multirow{2}{*}{AUC (\%)} \\
		\cline{2-5} \cline{7-10} \cline{12-15} 
		& 30\% & 20\% & 10\% & 1\% & 
		& 30\% & 20\% & 10\% & 1\% &
		& 30\% & 20\% & 10\% & 1\% &
		\\
		\hline \hline
		DeepID2 
		& {\color{black}12.5} & {\color{black}8.1} & {\color{black}3.3} & {\color{black}0.2}
		& {\color{black}20.8}
		& 4.0 & 2.1 & 0.8 & 0.1
		& 7.9
		& {\color{black}12.8} & {\color{black}8.1} & {\color{black}3.4} & {\color{black}0.8}
		& {\color{black}20.8}
		\\
		CentreFace 
		& \bf {\color{black}26.2} & \bf {\color{black}20.0} & \bf {\color{black}12.2} & \bf {\color{black}2.8}
		& \bf {\color{black}34.6}
		& 5.7 & 4.4 & 2.3 & 0.2
		& 7.6 
		& \bf {\color{black}27.3} & \bf {\color{black}21.0} & \bf {\color{black}13.8} & \bf {\color{black}3.1}
		& \bf {\color{black}37.3}
		\\
		FaceNet 
		& \color{black}10.6 & \color{black}7.9 & \color{black}3.6 & \color{black}0.5
		& \color{black}18.9
		& 4.0 & 3.0 & 0.8 & 0.1
		& 6.4
		& \color{black}12.7 & \color{black}8.1 & \color{black}4.3 & \color{black}1.0
		& \color{black}19.8
		\\
		VggFace 
		& - & - & - & - & - 	      
		& {\bf 6.5} & {\bf 4.8} & {\bf 2.5} & {\bf 0.2}
		& {\bf 9.6}
		& {\color{black}5.1} & {\color{black}2.6} & {\color{black}0.8} & {\color{black}0.1}
		& {\color{black}14.0}
		\\
		SphereFace 
		& {\color{black}18.8} & {\color{black}13.5} & {\color{black}7.0} & {\color{black}0.7}
		& {\color{black}26.6}
		& 5.9 & 4.2 & 2.2 & 1.7
		& 9.0
		& {\color{black}21.3} & {\color{black}15.7} & {\color{black}8.3} & {\color{black}1.0}
		& {\color{black}28.1}
		\\
		\hline 
	\end{tabular}
\end{table*}


\section{Experimental Results}\label{sec:FREvaluate}

In this section, we present and discuss the experimental evaluations of surveillance FR. 
The performances of varying FR methods are evaluated  
on both native low-resolution surveillance faces (Sec. \ref{sec:eval_LR_FR}) 
and super-resolved faces (Sec. \ref{sec:eval_SR_FR}).

\subsection{Low-Resolution Surveillance Face Recognition}
\label{sec:eval_LR_FR}

We evaluated the FR performance on the {\em native} 
low-resolution QMUL-SurvFace images with ambiguous observation.
Apart from the low-resolution issue, there are other uncontrolled
covariates and noises, e.g. illumination variations, expression, occlusions, background clutter, and compression artifacts.
All of these factors cause inference uncertainty to varying degrees
(Fig. \ref{fig:images_survface}).

\vspace{0.1cm}
\noindent {\bf Model Training and Test.}
For training a FR model, we adopted three strategies as below:
{\bf (1)} Only using QMUL-SurvFace training set (220,890 images from 5,319 IDs).
{\bf (2)} Only using CASIA web data (494,414 images from 10,575 IDs). 
We will test the effect of using different web source datasets 
such as MegaFace2 \citep{nech2017level} and MS-Celeb-1M~\citep{guo2016msceleb}.
{\bf (3)} First pre-training a FR model on CASIA, 
then 
fine-tuning on QMUL-SurvFace.
By default we adopt this training strategy. 
After a FR model is trained by any strategy as above, 
we deploy it with Euclidean distance.
%

%
%
In both training and test, 
we rescaled all facial images by {\em bicubic} interpolation
to the required size of a FR model.
%
Such interpolated images are still
of ``low-resolution'' since the underlying resolution is
mostly unchanged
(Fig. \ref{fig:SR_visualisation}).

\vspace{0.1cm}
\noindent {\bf Evaluation Settings.}
We considered face verification and identification.
By default, we adopt the more realistic open-set evaluation
for face identification,
unless stated otherwise.
For open-set test
(Sec. \ref{sec:exp_LR_Identification}), we used TPIR (Eqn. \eqref{eq:FNIR}) at 
varying FPIR rates (Eqn. \eqref{eq:FPIR}).
The true match ranked in top-$r$ (i.e. $r\!=\!20$ in Eqn. \eqref{eq:TPIR}) is considered
as success. 
For face verification test (Sec. \ref{sec:exp_LR_veri}),
we used TAR (Eqn. \eqref{eqn:TAR}) and FAR (Eqn. \eqref{eqn:FAR}).

\vspace{0.1cm}
\noindent {\bf Implementation Details.}
For CentreFace \citep{wen2016discriminative},
VggFace \citep{parkhi2015deep}, and
SphereFace \citep{liu2017sphereface}
we used the codes released by the original authors.
For FaceNet \citep{schroff2015facenet},
we utilised a TensorFlow reimplementation\footnote{\url{https://github.com/davidsandberg/facenet}}.
We reproduced DeepID2 \citep{sun2014deepid2}.
Throughout the experiments, 
we adopted the suggested parameter setting by the authors if available,
or carefully tuned the hyper-parameters by grid search.
\changed{Data augmentation was applied to
	QMUL-SurvFace training data, 
	including flipping, 
	Gaussian kernel blurring, colour shift, brightness and contrast adjustment.
We excluded cropping and rotation transformation 
which bring negative influence due to tight face bounding boxes.}


\subsubsection{Face Identification Evaluation}
\label{sec:exp_LR_Identification}

\begin{figure}[h] 
	\centering
	\includegraphics[width=.99\linewidth]{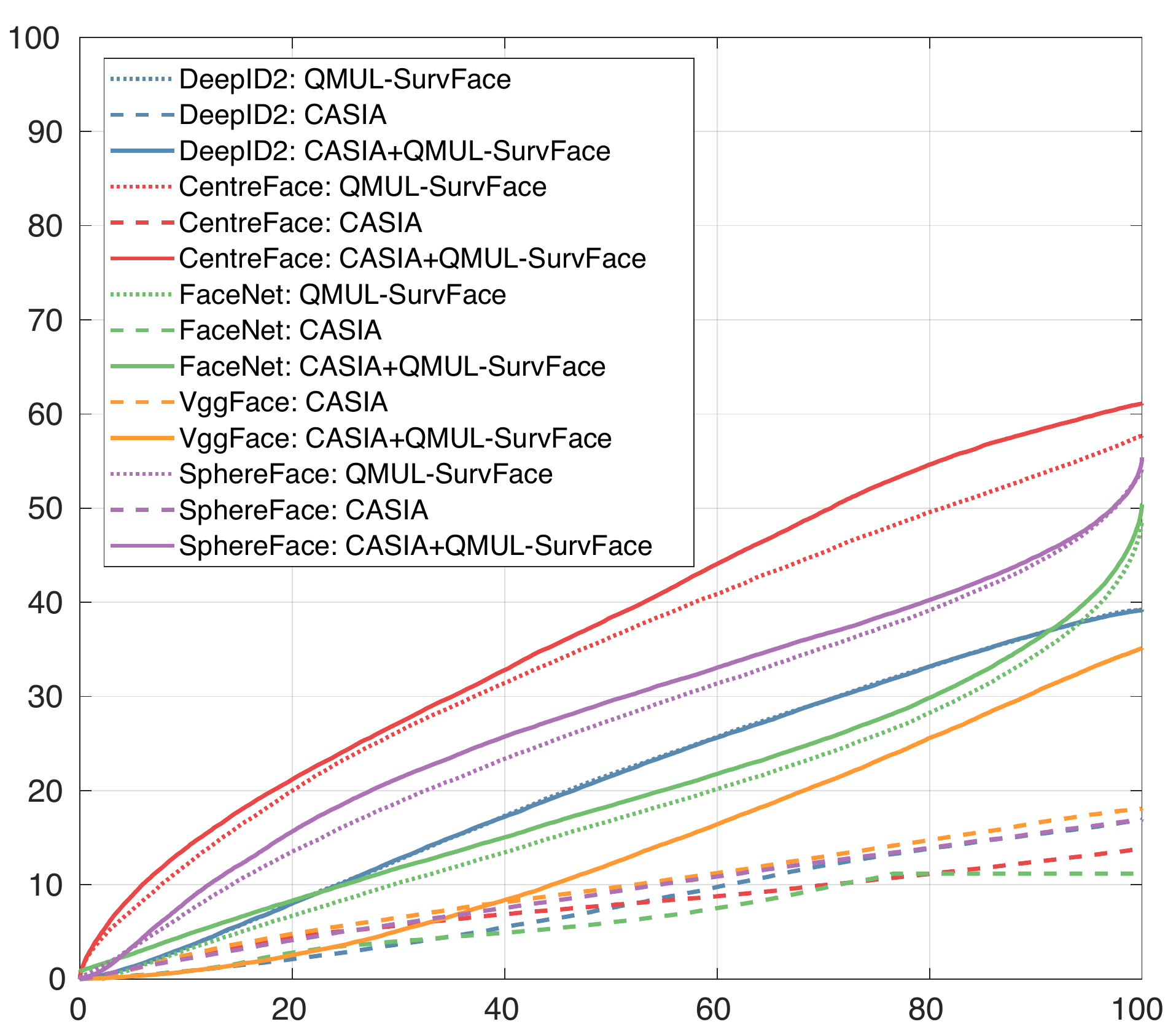}
	\vskip -0.2cm
	\caption{
		Face identification results on {\em QMUL-SurvFace}.
		Protocol: Open-Set.
		{Metrics}: TPIR$20$@FPIR ($r\!=\!20$).
	}
	\label{fig:LR_FR_curves}
\end{figure}

\begin{figure} 
	\centering
	\begin{subfigure}[t]{0.4\textwidth}
		\includegraphics[width=1\textwidth]{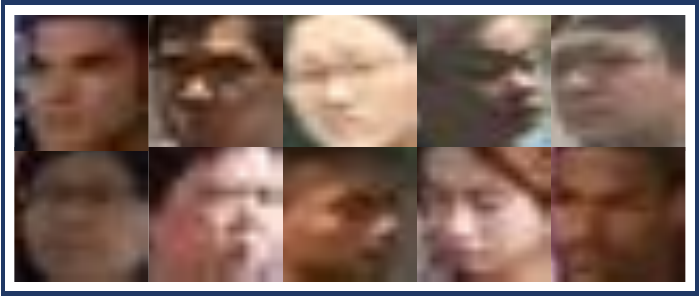}
		\caption{QMUL-SurvFace}
	\end{subfigure}
	\begin{subfigure}[t]{0.4\textwidth}
		\includegraphics[width=1\textwidth]{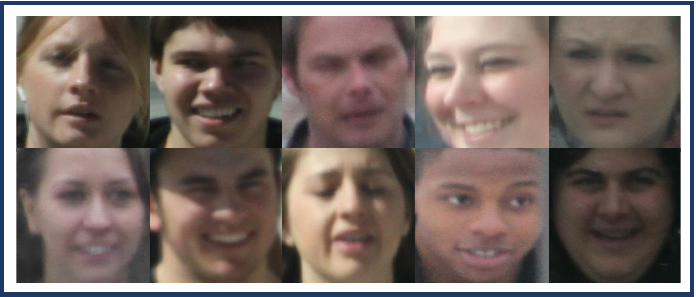}
		\caption{UCCS}
	\end{subfigure}
	\begin{subfigure}[t]{0.4\textwidth}
		\includegraphics[width=1\textwidth]{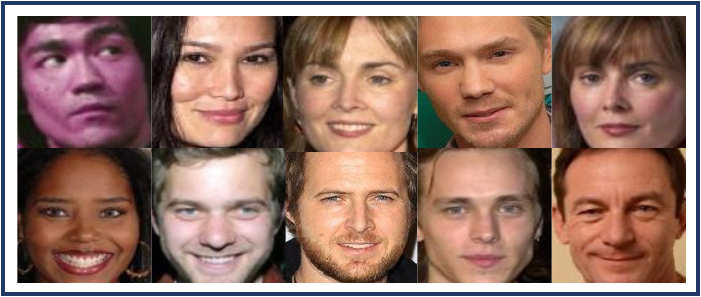}
		\caption{CASIA}
	\end{subfigure}
	\vskip -0.2cm
	\caption{
		Quality comparison of example faces from ({a}) QMUL-SurvFace, ({b}) UCCS, and ({c}) CASIA.
	}	
	\label{fig:UCCS-QMUL}
\end{figure}

\vspace{0.1cm}
\noindent {\bf (I) Benchmark QMUL-SurvFace.}
We benchmarked face verification on QMUL-SurvFace
in Table \ref{table:results_iden_open_survface} and Fig. \ref{fig:LR_FR_curves}. 
We make four observations:
\noindent {(1)} 
Not all FR models (e.g. VggFace)
converge when directly training on
QMUL-SurvFace. 
As opposite, all models are well trained with CASIA data.
Whilst the data size of CASIA 
is larger,
we conjugate that the scale is not a key obstacle 
as QMUL-SurvFace training data should be arguably sufficient
for generic deep model training.
Instead this may be more due to extreme challenges posed 
by poor resolution
especially when the model requires
high-scale inputs like $224\!\times\!224$ by VggFace.
This indicates the dramatic differences
between native surveillance FR and web FR.
\noindent {(2)} 
The poorest FR results are yielded by
the models trained with only CASIA faces.
This is expected
due to the clear domain gap between CASIA and QMUL-SurvFace
(Fig. \ref{fig:UCCS-QMUL}).
\noindent {(3)} 
Most models are notably improved once pre-trained using CASIA faces.
This suggests a positive effect of web data
by refining model initialisation.
\noindent {(4)} 
CentreFace is the best performer of the five models.
This indicates the efficacy of restricting intra-class variation in training
for surveillance FR,
consistent with web FR \citep{wen2016discriminative}.

\begin{table} [h]
	\centering
	\setlength{\tabcolsep}{0.4cm} 
	\caption{
		Open-Set vs Closed-Set 
		on {\em QMUL-SurvFace}.
	}
	\vskip -0.2cm
	\label{table:probe_effect_open_survface}
	\begin{tabular}{c|ccc}
		\hline
		\multirow{2}{*}{Metrics} & \multicolumn{3}{c}{TPIR$20$(\%)@FPIR (Open-Set)} 
		\\
		\cline{2-4}
		& 30\% & 20\% & 10\% 
		\\				 
		\hline \hline
		CentreFace & 27.3 & 21.0 & 13.8
		\\
		\hline
		\hline 
		SphereFace & 21.3 & 15.7 & 8.3
		\\
		\hline \hline
		\multirow{2}{*}{Metrics} & \multicolumn{3}{c}{CMC  (\%) (Closed-Set)} 
		\\
		\cline{2-4}
		& Rank-1 & Rank-10 & Rank-20 \\
		\hline
		CentreFace  & 29.9 & 53.4 & 61.1
		\\
		\hline
		SphereFace & 29.3 & 50.0 & 55.4
		\\
		\hline
	\end{tabular}
\end{table}

\vspace{0.1cm}
\noindent {\bf (II) Open-Set {\em vs} Closed-Set.}
We compared open-set FR with the conventional closed-set setting 
on QMUL-SurvFace.
In closed-set test, 
we removed all distractors in the test gallery.
We evaluated the top-2 FR models, CentreFace 
and SphereFace. %
Table \ref{table:probe_effect_open_survface} suggests that
{\em closed-set FR is clearly easier than the open-set counterpart}.
For instance, 
CentreFace achieves 27.3\% TPIR$20$@FPIR$30\%$ in open-set
{\em vs} 61.1\% Rank-20 in closed-set.
The gap is even larger at lower false alarm rates. 
This means that when taking into account the attacking of distractors, FR is much harder.

\vspace{0.1cm}
\noindent {\bf (III) SurvFace {\em vs} WebFace.}
We compared surveillance FR with web FR
in the closed-set test.
We used CentreFace for example.
This model achieves Rank-1 $29.9\%$ (Table \ref{table:probe_effect_open_survface}) on QMUL-SurvFace,
much inferior to the rate of $65.2\%$ on MegaFace, 
i.e. a $54\%$ (1-29.9/65.2) performance drop.
This indicates that surveillance FR is significantly more challenging,
especially so when considering that one million distractors are used 
to additionally complicate the MegaFace test.

\begin{table} [!h] 
	\centering
	\setlength{\extrarowheight}{0.1mm}
	\setlength{\tabcolsep}{0.1cm} 
	\caption{
		Image quality in face identification: UCCS {\em vs} QMUL-SurvFace(1090ID).
	}
	\vskip -0.2cm
	\label{table:eval_img_quality}
	\begin{tabular}{c|c||cccc|c}
		\hline  
		\multirow{2}{*}{Challenge} & \multirow{2}{*}{Model}
		& \multicolumn{4}{c|}{TPIR$20$(\%)@FPIR} & \multirow{2}{*}{AUC (\%)} \\
		\cline{3-6}
		& & 30\% & 20\% & 10\% & 1\% & 
		\\
		\hline \hline
		\multirow{5}{*}{UCCS}
		& DeepID2 
		& 70.7 & 61.7 & 48.3 & 10.0
		& 76.9 \\
		& CentreFace 
		& {\bf 96.1} & {\bf 94.6} & {\bf 90.4} & {\bf 80.7}
		& {\bf 96.1} \\
		& FaceNet 
		& 91.0 & 89.3 & 83.6 & 66.3
		& 91.5
		\\
		& VggFace 
		& 71.0 & 60.3 & 46.6 & 15.0
		& 77.0 \\
		& SphereFace 
		& 74.0 & 67.5 & 58.0 & 26.8
		& 76.5
		\\ \hline
		
		\multirow{5}{*}{QMUL}
		& DeepID2 
		& 45.0 & 34.0 & 22.0 & 4.6
		& 55.3
		\\
		& CentreFace
		& 52.0 & 46.0 & 35.0 & 13.0
		& 60.3
		\\
		& FaceNet
		& 55.0 & 49.3 & 40.5 & 16.0
		& 59.9
		\\
		& VggFace
		& 42.0 & 32.0 & 21.0 & 5.0
		& 51.0
		\\
		& SphereFace
		& \bf 59.9 & \bf 56.0 & \bf 49.0 & \bf 20.0
		& \bf 64.0
		\\
		\hline
	\end{tabular}
\end{table}

\vspace{0.1cm}
\noindent {\bf (IV) SurvFace Image Quality.}
We evaluated the effect of surveillance image quality in open-set FR. 
To this end,
we contrasted QMUL-SurvFace with UCCS \citep{gunther2017unconstrained} that 
provides surveillance face images with clearly better quality 
(Fig. \ref{fig:UCCS-QMUL}).

\vspace{0.1cm}
\noindent {\em Setting.}
For UCCS, we used the released face images from 1,090 out of 1,732 IDs
with the remaining not accessible. 
We made a 545/545 train/test ID random split,
resulting in a 6,948/7,068 image split.
To enable a like-for-like comparison, we constructed 
a 
{\em QMUL-SurvFace(1090ID)} dataset 
by randomly picking 545/545 QMUL-SurvFace train/test IDs.
For evaluation, we designed an open-set test setting using
100 random IDs 
for gallery and all 545 IDs
for probe.

\vspace{0.1cm}
\noindent {\em Results.} 
Table \ref{table:eval_img_quality} shows that 
QMUL-SurvFace poses more challenges than UCCS,
with varying degrees of performance drops experienced 
by different FR models.
This suggests that 
image quality is an important factor,
and UCCS is less accurate
in reflecting the surveillance FR challenges 
due to {\em artificially} high image quality. %

\begin{table} [!h]
	\centering
	\color{black}
	\setlength{\extrarowheight}{0.1mm}
	\setlength{\tabcolsep}{0.08cm} 
	\caption{
			LR-to-LR {\em vs} LR-to-HR FR on UCCS. 
	}
	\vskip -0.2cm
	\label{table:eval_hr2lr}
	\begin{tabular}{c|c||cccc|c}
		\hline 
		\multirow{2}{*}{Setting} & \multirow{2}{*}{Model}
		& \multicolumn{4}{c|}{TPIR$20$(\%)@FPIR} & \multirow{2}{*}{AUC (\%)} 
		\\ \cline{3-6} 
		& & 30\% & 20\% & 10\% & 1\% & \\ \hline \hline
		\multirow{2}{*}{LR-to-LR}
		& CentreFace 
		&\bf 92.0 &\bf 90.2 &\bf 87.0 & \bf 70.0
		&\bf 93.0 \\
		& FaceNet
		& 89.8 &\bf 87.5 &\bf 82.4 & \bf 55.3
		&\bf 90.9
		\\ \hline
		\multirow{2}{*}{LR-to-HR}
		& CentreFace 
		& 91.7 & 90.1 & 84.6 & 66.7
		& 92.8 \\
		& FaceNet
		& 89.8 & 86.6 & 79.1 & 50.3
		& 90.1
		\\
		\hline
	\end{tabular}
\end{table}

\vspace{0.1cm}
\noindent \changed{{\bf (V) LR-to-LR {\em vs} LR-to-HR.}
We compared the two common low-resolution FR deployment settings on
the UCCS benchmark dataset.
It is observed in Table~\ref{table:eval_hr2lr} that 
most performances are comparable,
suggesting that LR-to-LR and LR-to-HR exhibit similar 
low-resolution face recognition challenges.
Interestingly, LR-to-LR results are relatively better. This is due to
less resolution gap to bridge than those of LR-to-HR.}

\vspace{0.1cm}
\noindent {\bf (VI) Test Scalability.}
We examined the test scalability by comparing 
QMUL-SurvFace(1090ID) and QMUL-SurvFace.
It is shown in the comparison between
Table \ref{table:results_iden_open_survface} and
Table \ref{table:eval_img_quality}
that 
significantly higher FR performances are yielded 
on the smaller test with 1090 IDs. 
%
This suggests that a large test benchmark
is necessary and crucial for enabling 
the true performance evaluation of practical applications.

\begin{table} [!h]
	\centering
	\color{black}
	\renewcommand{\arraystretch}{1}
	\setlength{\tabcolsep}{0.1cm}
	\caption{Effect of web image resolution. 
	}
	\vskip -0.3cm
	\begin{tabular}{c|c||cccc|c}
		\hline 
		\multirow{2}{*}{Resolution} & 
		\multirow{2}{*}{Model}
		& \multicolumn{4}{c|}{TPIR$20$(\%)@FPIR} & \multirow{2}{*}{AUC (\%)} \\
		\cline{3-6} 
		& & 30\% & 20\% & 10\% & 1\% & 
		\\ \hline \hline
		\multirow{5}{*}{24$\times$20} 
		& DeepID2 
		& {\color{black}12.7} & {\color{black}8.1} & {\color{black}3.4}  & {\color{black}0.2}
		& {\color{black}20.6}
		\\
		& CentreFace 
		& {\color{black}27.0} & {\color{black}21.0} &\bf {\color{black}14.0} & {\bf \color{black}3.2}
		& {\color{black}37.3} \\
		& FaceNet 
		& \color{black}12.3 & \color{black}8.0 & \color{black}4.3 & \color{black}0.5
		& \color{black}19.6
		\\
		& VggFace 
		& {\color{black}5.1} &\bf {\color{black}2.8} &\bf {\color{black}1.1} & {\color{black}0.1}
		& {\color{black}11.3}
		\\
		
		& SphereFace 
		& {\color{black}-} & {\color{black}-} & {\color{black}-} & {\color{black}-} 
		& {\color{black}-}
		\\
		\hline
		
		\multirow{5}{*}{{\color{black} 112$\times$96}}
		& DeepID2
		& \bf{\color{black}12.8} & {\color{black}8.1} & {\color{black}3.4} &\bf {\color{black}0.8}
		&\bf {\color{black}20.8}
		\\
		
		& CentreFace
		&\bf {\color{black}27.3} & {\color{black}21.0} & {\color{black}13.8} & {\color{black}3.1}
		& {\color{black}37.3}
		\\
		& FaceNet 
		& \bf \color{black}12.7 & \bf \color{black}8.1 & \color{black}4.3 & \bf \color{black}1.0
		& \bf \color{black}19.8
		\\
		& VggFace
		& {\color{black}5.1} & {\color{black}2.6} & {\color{black}0.8} & {\color{black}0.1}
		&\bf {\color{black}14.0}
		\\
		& SphereFace
		&\bf {\color{black}21.3} &\bf {\color{black}15.7} &\bf {\color{black}8.3} &\bf {\color{black}1.0}
		&\bf {\color{black}28.1}
		\\
		\hline 
	\end{tabular}
	\label{tab:web_data_res}
\end{table}

\vspace{0.1cm}
\noindent \changed{
	{\bf (VII) Web Image Resolution.}
	We examined the impact of CASIA web face resolution on surveillance FR.
	We compared the common 112$\times$96 size 
	with QMUL-SurvFace average resolution 24$\times$20.
	It is observed in Table \ref{tab:web_data_res} that 
	matching web face to QMUL-SurvFace in resolution 
	does not bring performance benefit in most cases.
	This is because
	web face images only give limited contribution
	for surveillance FR 
	due to severe domain discrepancy (Table \ref{table:results_iden_open_survface}),
	and simply aligning image resolution cannot 
	alleviate this problem.
}

\begin{table} [!h]
	\centering
	\color{black}
	\renewcommand{\arraystretch}{1}
	\setlength{\tabcolsep}{0.09 cm}
	\caption{Selection of web training data source.
	}
	\vskip -0.3cm
	\begin{tabular}{c|c||cccc|c}
		\hline 
		\multirow{2}{*}{Web Dataset} &
		\multirow{2}{*}{Model}
		& \multicolumn{4}{c|}{TPIR$20$(\%)@FPIR} & \multirow{2}{*}{AUC (\%)} \\
		\cline{3-6} 
		& & 30\% & 20\% & 10\% & 1\% &
		\\ \hline \hline
		\multirow{2}{*}{MS-Celeb-1M}
		& CentreFace 
		& {\bf \color{black}28.0} & {\bf \color{black}21.9} & {\color{black}14.1} & {\color{black}3.1}
		& {\color{black}37.4} 
		\\
		& SphereFace 
		& 20.1 & 13.6 & 5.4 & 0.8
		& 27.1
		\\ \hline
		
		\multirow{2}{*}{MegaFace2}
		& CentreFace
		& {\color{black}27.7} & {\bf \color{black}21.9} & {\bf \color{black}15.0} & {\bf \color{black}3.5}
		& {\bf \color{black}37.6}
		\\
		& SphereFace
		& 20.0 & 13.0 & 5.4 & 0.7
		& 26.4 
		\\ \hline
		
		\multirow{2}{*}{CASIA}
		& CentreFace
		& {\color{black}27.3} & {\color{black}21.0} & {\color{black}13.8} & {\color{black}3.1}
		& {\color{black}37.3}
		\\ 
		& SphereFace
		& {\bf \color{black}21.3} & {\bf \color{black}15.7} & {\bf \color{black}8.3} & {\bf \color{black}1.0}
		& {\bf \color{black}28.1}
		\\
		\hline 
	\end{tabular}
	\label{tab:web_data}
\end{table}

\vspace{0.1cm}
\noindent \changed{
{\bf (VIII) Web Image Source.}
We tested the performance effect of web training dataset 
by comparing three benchmarks:
CASIA \citep{yi2014learning}, MS-Celeb-1M~\citep{guo2016msceleb}, 
and MegaFace2~\citep{nech2017level}.
We evaluated CentreFace and SphereFace optimised 
by the suggested training strategies.
Table \ref{tab:web_data} shows that 
the selection of web data only leads to neglectable changes in surveillance FR
performance.
For training, CASIA is tens of times 
more cost-effective (cheaper) than the other two larger datasets,
so we use it
in the main experiments.
Moreover, it is more challenging to train a FR model 
given a vast ID class space such as MegaFace2.
}

\begin{table} [h] 
	\centering
	\color{black}
	\renewcommand{\arraystretch}{1}
	\setlength{\tabcolsep}{0.1 cm}
	\caption{Effect of QMUL-SurvFace image resolution. 
	}
	\vskip -0.3cm
	\begin{tabular}{c|c||cccc|c}
		\hline 
		{Face Width} &
		\multirow{2}{*}{Model}
		& \multicolumn{4}{c|}{TPIR$20$(\%)@FPIR} & \multirow{2}{*}{AUC (\%)} \\
		\cline{3-6} 
		(Pixels)& & 30\% & 20\% & 10\% & 1\% & 
		\\ \hline \hline
		
		\multirow{2}{*}{$\leq$20}
		& CentreFace 
		&\bf 32.9 &\bf 23.3 &\bf 15.2 &\bf 4.0
		&\bf 40.0
		\\
		& SphereFace 
		& 13.9 & 9.4 & 4.3 & 0.6
		& 22.4
		\\\hline
		
		\multirow{2}{*}{$>$20}
		& CentreFace
		& 25.0 & 19.7 & 13.5 & 3.4
		& 34.6
		\\
		& SphereFace
		&\bf 26.2 &\bf 21.6 &\bf 14.8 &\bf 2.7
		&\bf 32.2
		\\ \hline
		
		\multirow{2}{*}{All}
		& CentreFace
		& 27.3 & 21.0 & 13.8 & 3.1
		& 37.3
		\\
		& SphereFace
		& 21.3 & 15.7 & 8.3 & 1.0
		& 28.1
		\\
		\hline 
	\end{tabular}
	\label{tab:face_scale}
\end{table}

\vspace{0.1cm}
\noindent \changed{
{\bf (IX) SurvFace Resolution.}
We examined the effect of test image resolution.
Given the bi-modal distribution of QMUL-SurvFace images,
we divided all test probe faces into 
two groups at the threshold of 20 pixels in width
and evaluated the FR performance on each group.
Table \ref{tab:face_scale} shows that 
whilst the face dimension matters, 
the average performance on all test images 
summarises rather well that of each group.
The performance variation across groups relies on both FR models applied
and other imaging factors, 
suggesting that the resolution {\em alone}
does not bring a {\em consistent} performance bias.
}

\begin{figure} [h]
	\centering
	\begin{subfigure}[t]{0.45\textwidth}
		\includegraphics[width=1\textwidth]{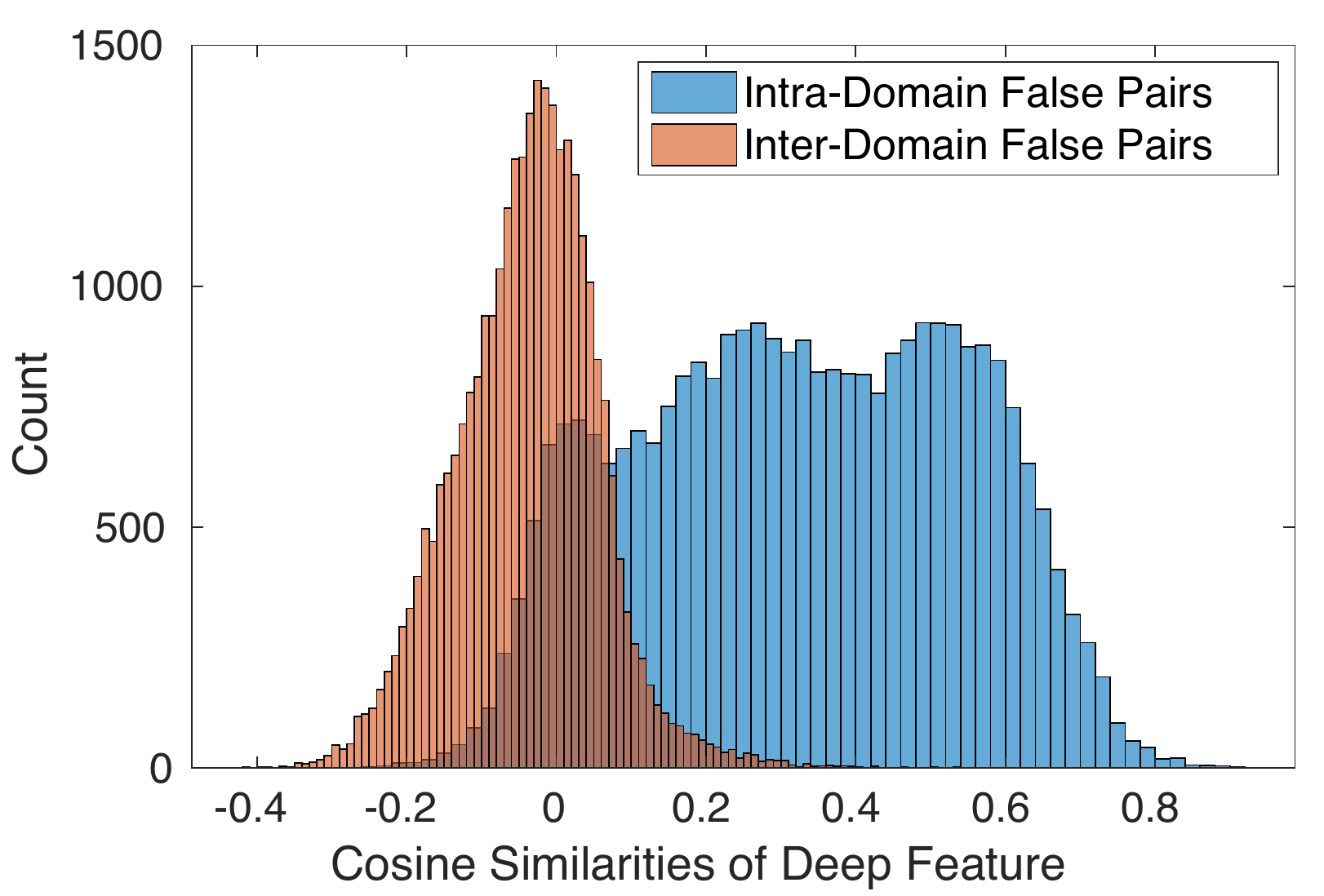}
		\caption{In the {\em CentreFace} feature space.}
	\end{subfigure}
	\vskip 0.3cm
	\begin{subfigure}[t]{0.45\textwidth}
		\includegraphics[width=1\textwidth]{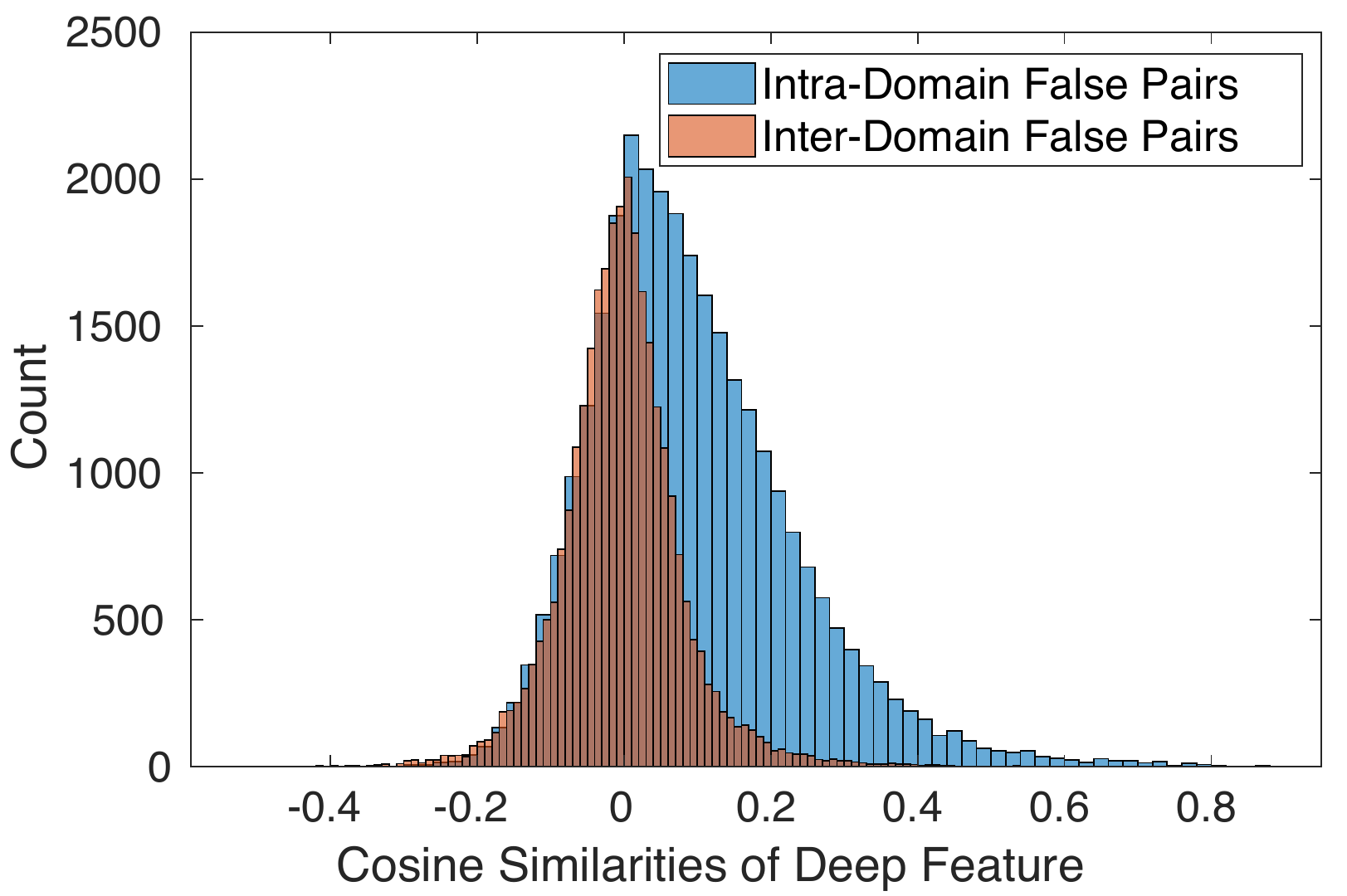}
		\caption{In the {\em SphereFace} feature space.}
	\end{subfigure}
	\caption{
		Similarity distributions 
		of {\em Intra-Domain} and {\em Inter-Domain} false pairs in the feature space 
		established by (a) CentreFace and (b) SphereFace.
	}
	\label{fig:domain_seg}
\end{figure}

\vspace{0.1cm}
\noindent \changed{{\bf (X) Intra-Domain Similarity Effect.}
We analysed the intra-domain similarity effect in QMUL-SurvFace -- 
That is, we
tested whether face images from one source (domain) are all easily 
different from other domains.
This common domain
characteristics may overwhelm the more subtle facial identity
differences across domains. 
%
To that end, 
we examined the similarity statistics of {\em intra-domain} 
and {\em inter-domain} false pairs.
Specifically, we formed 60,000 intra-domain and 60,000 inter-domain probe-gallery
false pairs, 
and profiled their cosine similarity values
measured by the CentreFace and SphereFace features, respectively. 
Fig.~\ref{fig:domain_seg} shows that
the intra-domain similarity effect is not severe
with 
a large overlap between inter-domain and intra-domain
in the matching score (x-axis) range for both CentreFace and SphereFace. 
This indicates that different source domains of QMUL-SurvFace
do not impose implicitly trivial dominant intra-domain influence,
therefore this benchmark does provide a meaningfully challenging
open-set test. 
}

\begin{figure*} [h]
	\centering
	\includegraphics[width=1\textwidth]{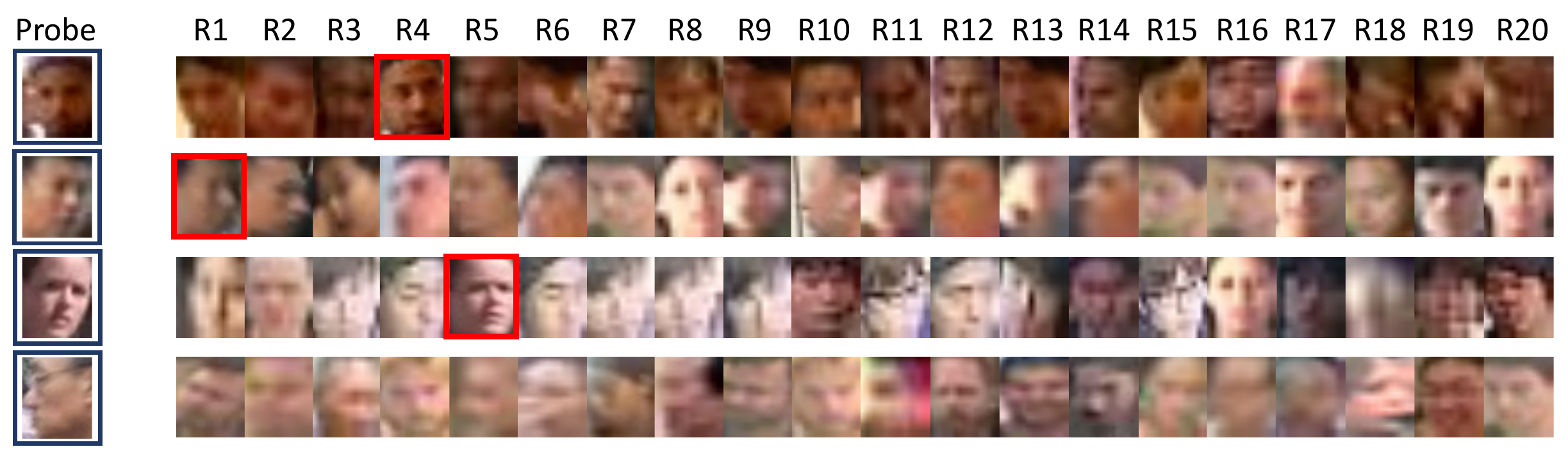}
	\vskip -0.3cm
	\caption{
		Face identification examples by CentreFace on 
		{\em QMUL-SurvFace}.
		True matches are in red box.
	}
	\label{fig:retrieval_results}
\end{figure*}

\vspace{0.1cm}
\noindent {\bf (XI) Qualitative Evaluation.}
To provide a visual evaluation, we show 
face identification examples by CentreFace on QMUL-SurvFace in Fig. \ref{fig:retrieval_results}.
The model succeeds in finding the true match among top-20 in the top three tasks,
and fails the last.
Poor quality surveillance data present extreme FR challenges.

\subsubsection{Face Verification Evaluation}
\label{sec:exp_LR_veri}

\begin{table} [h]
	\centering
	\setlength{\extrarowheight}{0.1mm}
	\setlength{\tabcolsep}{0.25cm} 
	\caption{Face verification results on {\em QMUL-SurvFace}. 
	}
	\vskip -0.2cm
	\label{table:results_ver_full}
	\begin{tabular}{c|cccc|c}
		\hline 
		\multirow{2}{*}{Model}
		& \multicolumn{4}{c|}{TAR(\%)@FAR} & \multirow{2}{*}{AUC (\%)} \\
		\cline{2-5} 
		& 30\% & 10\% & 1\% & 0.1\% & \\
		\hline \hline
		DeepID2  		      
		& 80.6 & 60.0 & 28.2 & 13.4 & 84.1 \\
		{CentreFace }  
		& \bf 95.2 & \bf 86.0 &  \bf 53.3 & \bf 26.8 & \bf 94.8 \\
		FaceNet 		  
		& 94.6 & 79.9 & 40.3 & 12.7 & 93.5
		\\
		VggFace 		      
		& 83.2 & 63.0 & 20.1 & 4.0 & 85.0
		\\
		SphereFace  
		& 80.0 & 63.6 & 34.1 & 15.6 & 83.9
		\\
		\hline
	\end{tabular}
\end{table}

\vspace{0.1cm}
\noindent {\bf (I) Benchmark QMUL-SurvFace.}
We benchmarked face verification on QMUL-SurvFace.
It is shown in Table \ref{table:results_ver_full},
that CentreFace 
remains the best performer as in face identification (Table \ref{table:results_iden_open_survface}). 
A divergence is that FaceNet 
achieves the $2^\text{nd}$ position, beating SphereFace.
This suggests an operational difference
between face identification and verification.
%
All models perform poorly at low FAR (e.g. 0.1\%),
indicating that face verification in surveillance images 
is still an unsolved task. 

\begin{table} [h] 
	\centering
	\setlength{\extrarowheight}{0.1mm}
	\setlength{\tabcolsep}{0.9cm} 
	\caption{Face verification results on {\em QMUL-SurvFace}.
	}
	\vskip -0.2cm
	\label{table:results_ver_mean_acc_full}
	\begin{tabular}{c|c}
		\hline
		Model & Mean Accuracy (\%) \\
		\hline \hline
		DeepID2 
		& 76.12  \\
		CentreFace 
		& \bf 88.00  \\
		FaceNet 
		& 85.31 \\
		VggFace 
		& 77.96 \\
		SphereFace 
		& 77.57  \\
		\hline  
	\end{tabular}
\end{table}

\vspace{0.1cm}
\noindent {\bf (II) SurvFace {\em vs} WebFace.}
We used LFW \citep{huang2007labeled} to compare surveillance and web image
based face verification.
To this end, 
we additionally evaluated QMUL-SurvFace by 
{\em mean accuracy} (as Table \ref{table:LFW}), 
i.e. the fraction of correctly verified pairs.
%
It is seen in Table \ref{table:results_ver_mean_acc_full} 
that the best performance on 
QMUL-SurvFace
is $88.00\%$ by CentreFace, 
considerably lower
than the best LFW result $99.83\%$.
This suggests a higher challenge of FR in surveillance
imagery data.

\begin{table} [h]
	\centering
	\setlength{\extrarowheight}{0.1mm}
	\setlength{\tabcolsep}{0.1cm} 
	\caption{Image quality in face verification: UCCS {\em vs} QMUL-SurvFace(1090ID). 
	}
	\vskip -0.2cm
	\label{table:results_ver}
	\begin{tabular}{c|c|cccc|c}
		\hline 
		\multirow{2}{*}{Challenge} 
		& 
		\multirow{2}{*}{Model}
		& \multicolumn{4}{c|}{TAR(\%)@FAR} & \multirow{2}{*}{AUC (\%)}\\
		\cline{3-6}
		& & 30\% & 10\% & 1\% & 0.1\% & \\
		\hline \hline
		\multirow{5}{*}{UCCS}
		& DeepID2 
		& 93.1 & 83.4 & 61.7 & 37.9 & 93.8
		\\
		& CentreFace 
		& \bf 99.6 & \bf 97.0 &  \bf 87.8 & \bf 75.5 & \bf 99.0
		\\
		& FaceNet 
		& 98.2 & 93.8 & 79.4 & 63.4 & 97.8
		\\
		& VggFace 
		& 97.1 & 90.6 & 72.4 & 55.1 & 96.7
		\\
		& SphereFace 
		& 94.0 & 84.9 & 60.2 & 24.7 & 94.1
		\\
		\hline

		\multirow{5}{*}{QMUL}
		& DeepID2
		& 77.3 & 56.9 & 24.3 & 13.1 & 82.5
		\\
		& CentreFace
		& \bf 87.6 & \bf 69.5 & 31.1 & 9.5 & \bf 88.5 
		\\
		& FaceNet 
		& 82.9 & 61.3 & 29.2 & \bf 12.5 & 84.6
		\\
		& VggFace 
		& 77.4 & 54.6 & 23.4 & 8.9 & 82.1
		\\
		& SphereFace
		& 80.2 & 62.3 & \bf 32.5 & \bf 12.5 & 84.1
		\\
		\hline
	\end{tabular}
\end{table}

\vspace{0.1cm}
\noindent {\bf (III) SurvFace Image Quality.}
We evaluated the effect of surveillance image quality in face verification
by contrasting 
QMUL-SurvFace(1090ID) and UCCS.
We similarly generated 5,319 positive and 5,319 negative test UCCS pairs
(Table \ref{table:survface_veri}).
For a fair comparison,
we used only the 7,922 training images of 545 IDs in QMUL-SurvFace(1090ID)
for model training.
It is evident in Table \ref{table:results_ver} that
the FR performance
on UCCS is clearly higher than on QMUL-SurvFace.
%
This is consistent with face identification test 
(Table \ref{table:eval_img_quality}).

\begin{figure}[h] 
	\centering
	\begin{subfigure}{0.48\textwidth}
		\includegraphics[width=1\textwidth]{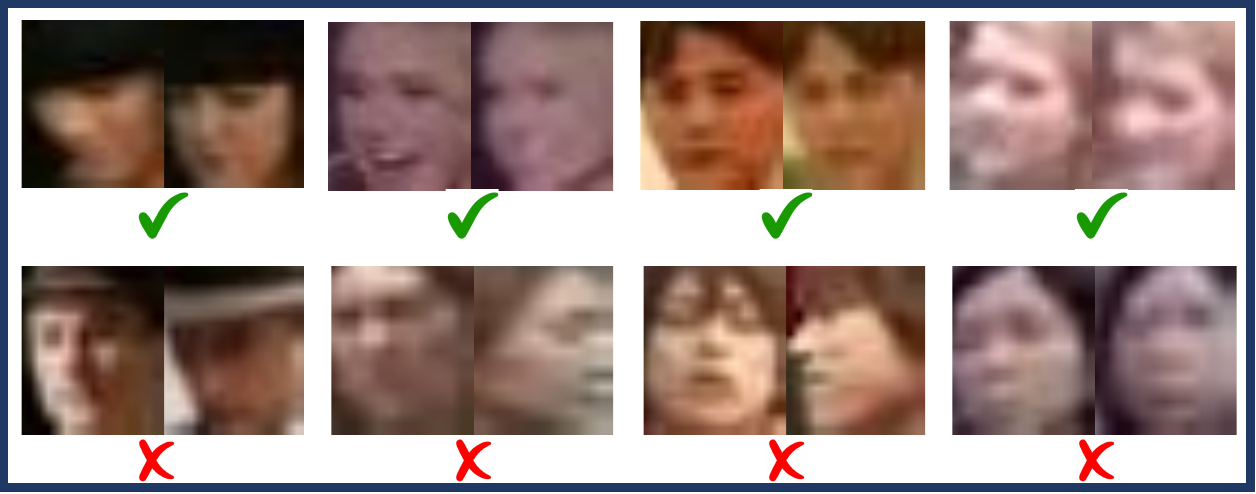}
		\caption{Ground-truth: {\em matched} pairs.}
	\end{subfigure}
	\begin{subfigure}{0.48\textwidth}
		\includegraphics[width=1\textwidth]{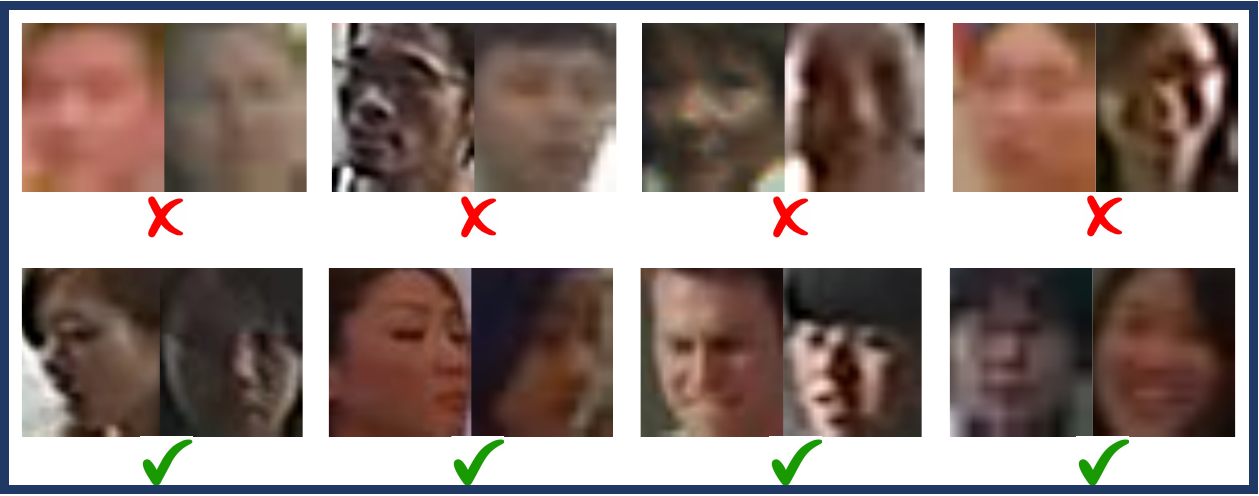}
		\caption{Ground-truth: {\em unmatched} pairs.}
	\end{subfigure}
	\vskip -0.2cm
	\caption{
	Face verification examples by CentreFace on {\em QMUL-SurvFace}.
		Failure cases are indicated in red.
	}	
	\label{fig:verfication_visual}
\end{figure}

\vspace{0.1cm}
\noindent {\bf (IV) Qualitative Evaluation.}
To give visual examination, 
we show face verification examples 
by CentreFace at FAR=10\% on QMUL-SurvFace in Fig. \ref{fig:verfication_visual}.


\begin{table*} [h] 
	\centering
	\setlength{\extrarowheight}{0.1mm}
	\setlength{\tabcolsep}{0.1cm} 
	\caption{
		Effect of image super-resolution in face identification on
		{\em  QMUL-SurvFace}.
		{Protocol}: Open-Set.
		SR: Super-Resolution.
		Jnt/Ind Train: Joint/Independent Training.
	}
	\vskip -0.2cm
	\label{table:SRFR_open}
	\begin{tabular}{l|c||cccc|c||cccc|c||cccc|c}
		\hline
		\multicolumn{2}{c||}{\multirow{2}{*}{Metrics}}
		& \multicolumn{4}{c|}{TPIR$20$(\%)@FPIR} & \multirow{2}{*}{AUC(\%)}
		& \multicolumn{4}{c|}{TPIR$20$(\%)@FPIR} & \multirow{2}{*}{AUC(\%)}
		& \multicolumn{4}{c|}{TPIR$20$(\%)@FPIR} & \multirow{2}{*}{AUC(\%)}
		\\
		\cline{3-6} \cline{8-11} \cline{13-16} 
		\multicolumn{2}{c||}{}
		& 30\% & 20\% & 10\% & 1\% & 
		& 30\% & 20\% & 10\% & 1\% & 
		& 30\% & 20\% & 10\% & 1\% & 
		\\
		\hline \hline 
		\multicolumn{2}{c||}{\backslashbox{SR Model \kern 0em}{FR Model}}
		& \multicolumn{5}{c||}{DeepID2}
		& \multicolumn{5}{c||}{CentreFace } 
		& \multicolumn{5}{c}{FaceNet}   
		\\ 
		\hline 
		\multicolumn{2}{c||}{No SR}   
		& {\bf \color{purple}12.8} & {\bf \color{purple}8.1} & {\bf \color{purple}3.4} & {\bf \color{purple}0.8}
		& {\bf \color{purple}20.8}
		& {\bf \color{purple}27.3} & {\bf \color{purple}21.0} & {\bf \color{purple}13.8} & {\bf \color{purple}3.1}
		& {\bf \color{purple}37.3}
		& \color{purple}\bf 12.7 & \color{purple}\bf 8.1 & \color{purple}\bf 4.3 & \color{purple}\bf 1.0
		& \color{purple}\bf 19.8
		\\
		\hline 
		\multirow{2}{*}{SRCNN}  
		& Ind Train 
		& \bf 12.3 & \bf 8.0 & \bf 3.4 & \bf 0.3
		& \bf 19.8
		& 25.0 & {20.0} & \bf {13.1} & \bf 3.0
		& 35.0
		& 12.3 & 8.0 & 4.1 & 0.5
		& 19.6
		\\
		& Jnt Train 
		& 6.0 & 3.7 & 1.6 & 0.2
		& 8.5
		& \bf {25.5} & \bf {20.5} & 12.0 & 2.9
		& 35.0
		& - & - & - & - 
		\\
		\hline 
		\multirow{2}{*}{FSRCNN} 
		& Ind Train 
		& {\bf 11.0} & {\bf 7.5} & {\bf 3.1} & {0.1}
		& {\bf 19.5}
		& {25.0} & \bf {20.0} & \bf {12.9} & {\bf 3.0}
		& \bf {35.0}
		& 12.0 & 7.8 & 4.0 & 0.5
		& 19.6
		\\
		& Jnt Train 
		& {8.2} & {4.9} & {2.1} & {\bf0.2}
		& {15.8}
		& {25.0} & {19.0} & {11.1} & {2.9}
		& {32.8}
		& - & - & - & - & -
		\\
		\hline 
		\multirow{2}{*}{VDSR} 
		& Ind Train 
		&\bf 12.2 & {\bf7.0} & {\bf3.3} & {0.2}
		& {\bf19.9}
		& {25.5} & {20.1} & {\bf12.8} & 3.0
		& {35.1}
		& 12.1 & 7.8 & 4.0 & 0.6
		& 19.7
		\\
		& Jnt Train 
		& {9.5} & {5.7} & {2.4} &\bf 0.3
		& {18.1}
		& {\bf26.7} & {\bf20.4} & {12.6} &\bf 3.1
		& \bf35.3
		& - & - & - & - & -
		\\
		\hline
		\multirow{2}{*}{DRRN}  
		& Ind Train 
		& 12.1 & {7.7} & {3.2} & {0.2}
		& {19.9}
		& {25.5} & {20.6} & {12.5} & 3.0
		& {35.0}
		& 12.3 & 7.9 & 4.1 & 0.6
		& 19.5
		\\
		& Jnt Train 
		& - & - & - & - & - 
		& - & - & - & - & - 
		& - & - & - & - & - 
		\\
		\hline
		\multirow{2}{*}{LapSRN}  
		& Ind Train 
		& {12.0} & {7.8} & {3.1} & {0.2}
		& {19.9}
		& {25.6} & {20.0} & {12.7} & {3.0}
		& {35.1}
		& 12.3 & 7.9 & 4.1 & 0.5
		& 19.6
		\\
		& Jnt Train 
		& - & - & - & - & - 
		& - & - & - & - & - 
		& - & - & - & - & -
		\\
		\hline \hline
		\multicolumn{2}{c||}{\backslashbox{SR Model \kern 0em}{FR Model}}
		& \multicolumn{5}{c||}{VggFace} 
		& \multicolumn{5}{c||}{SphereFace} \\
		\cline{1-12} 
		\multicolumn{2}{c||}{No SR}
		& { 5.1} & { 2.6} & { 0.8} & { 0.1}
		& { 14.0}
		& \bf \color{purple}{ 21.3} & \bf \color{purple}{ 15.7} & \bf \color{purple}{ 8.3} & \bf \color{purple}{ 1.0}
		& \bf \color{purple}{ 28.1}
		\\
		\cline{1-12} 
		\multirow{2}{*}{SRCNN}  
		& Ind Train 
		& \bf \color{purple}{ 6.2} & \bf \color{purple}{ 3.1} & \bf \color{purple}{ 1.0} & \bf \color{purple}{ 0.1}
		& \bf \color{purple}{ 15.3}
		& { 20.0} & { 14.9} & { 6.2} & { 0.6}
		& { 27.0}
		\\
		& Jnt Train 
		& - & - & - & - & -
		& - & - & - & - & -
		\\
		\cline{1-12} 
		\multirow{2}{*}{FSRCNN} 
		& Ind Train 
		& { 5.4} & { 2.7} & { 0.8} & { 0.1}
		& { 14.3}
		& { 20.0} & { 14.4} & { 6.1} & { 0.7}
		& { 27.3}
		\\
		& Jnt Train 
		& - & - & - & - & -
		& - & - & - & - & -
		\\
		\cline{1-12} 
		\multirow{2}{*}{VDSR} 
		& Ind Train 
		& { 5.8} & { 2.9} & { 1.0} & { 0.1}
		& { 15.0}
		& { 20.1} & { 14.5} & { 6.1} & { 0.8}
		& { 27.3}
		\\
		& Jnt Train 
		& - & - & - & - & - 
		& - & - & - & - & -
		\\
		\cline{1-12} 
		\multirow{2}{*}{DRRN}  
		& Ind Train 
		& { 5.8} & { 2.9} & { 0.8} & { 0.1}
		& { 15.1}
		& { 20.3} & { 14.9} & { 6.3} & { 0.6}
		& { 27.5} 
		\\
		& Jnt Train 
		& - & - & - & - & - 
		& - & - & - & - & -
		\\
		\cline{1-12} 
		\multirow{2}{*}{LapSRN}  
		& Ind Train 
		& { 5.7} & { 2.8} & { 0.9} & { 0.1}
		& { 15.0}
		& { 20.2} & { 14.7} & { 6.3} & { 0.7}
		& { 27.4}
		\\
		& Jnt Train 
		& - & - & - & - & -
		& - & - & - & - & -
		\\
		\cline{1-12} 
	\end{tabular}
\end{table*}

\subsection{Super-Resolution in Surveillance Face Recognition}\label{sec:eval_SR_FR}
Following the FR evaluation in raw low resolution surveillance data,
we tested {\em super-resolved} 
face images.
The aim is to examine the effect of image super-resolution or hallucination
techniques
in addressing the low-resolution problem in surveillance FR.
In this test, 
we only evaluated the native low-resolution QMUL-SurvFace benchmark,
since UCCS images are {\em artificially} of high resolution therefore excluded (Fig. \ref{fig:UCCS-QMUL}). 


\vspace{0.1cm}
\noindent {\bf Model Training and Test.}
For enhancing deep FR models with image super-resolution capability, 
we consider two training strategies.

\vspace{0.1cm}
\noindent{(1)} \textbf{\em Independent Training}: 
We first pre-train FR models 
on CASIA \citep{liu2015faceattributes} and 
then fine-tune on QMUL-SurvFace,
same as in Sec.
\ref{sec:eval_LR_FR}.
We subsequently {\em independently} train image super-resolution models with
CASIA data alone,
since no high-resolution QMUL-SurvFace data are available.
Low-res CASIA faces are generated by
down-sampling high-resolution ones, which together 
form training pairs for super-resolution models.
In test, we deployed the learned super-resolution models to 
restore low-resolution QMUL-SurvFace images
before performing FR by deep features and Euclidean distance.

\vspace{0.1cm}
\noindent{(2)} \textbf{\em Joint Training}:
Training super-resolution and FR models {\em jointly} in a hybrid pipeline
to improve their compatibility.
Specifically, we unit super-resolution and FR models by 
connecting the former's output with the latter's input 
so allowing an end-to-end training of both.
In practice, we first performed joint learning with CASIA
%
and then fine-tuned FR with QMUL-SurvFace. 
%
But joint training is not always
feasible 
due to additional challenges such as over-large model size and 
more challenging to converge the model training.
In our experiments, we achieved joint training of
six hybrid pipelines among three super-resolution
(SRCNN \citep{dong2014learning}, FSRCNN \citep{dong2016accelerating}, VDSR \citep{kim2016accurate}) 
and two FR models (DeepID2 \citep{sun2014deepid2} and CentreFace \citep{wen2016discriminative}).
At test time, we deployed the hybrid pipeline on QMUL-SurvFace images
to perform FR using Euclidean distance.

\vspace{0.1cm}
\noindent {\bf Evaluation Settings.}
For performance metrics, we used 
TPIR (Eqn. \eqref{eq:FNIR}) and
FPIR (Eqn. \eqref{eq:FPIR})
for face verification
and 
TAR (Eqn. \eqref{eqn:TAR}) and FAR (Eqn. \eqref{eqn:FAR})
for face identification,
same as Sec. \ref{sec:eval_LR_FR}.

\vspace{0.1cm}
\noindent {\bf Implementation Details.}
For super-resolution, 
we performed a 4$\times$ upscaling restoration.
We used the released codes by the authors 
for all super-resolution models.
In model training, we followed the parameter setting as suggested by the authors if available,
or carefully tuned them throughout the experiments.

\begin{figure*} 
	\centering
	\includegraphics[width=1\textwidth]{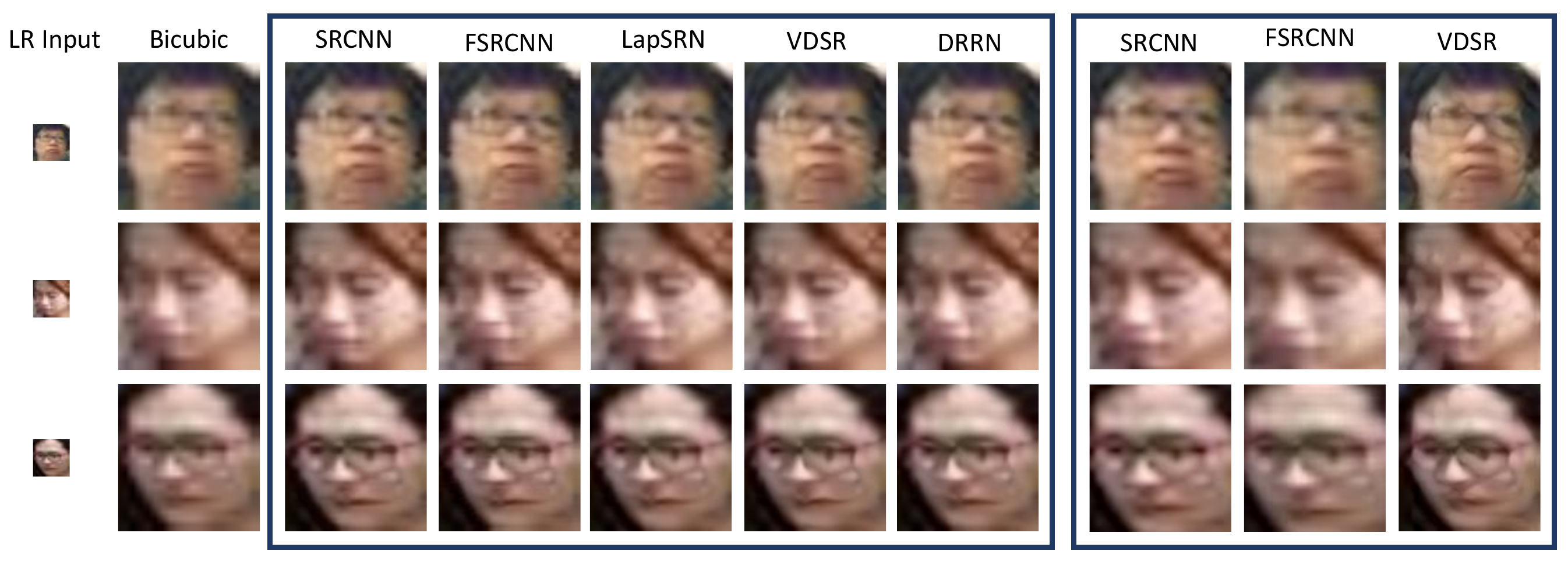}
	\vskip -0.2cm
	\caption{Super-resolved images on QMUL-SurvFace
	by independently ({\bf left box}) 
	and jointly trained ({\bf right box}) super-resolution models.
	CentreFace is used
	 in joint training.
}
	\label{fig:SR_visualisation}
\end{figure*}

\subsubsection{Face Identification Evaluation}

\vspace{0.1cm}
\noindent {\bf (I) Effect of Super-Resolution.}
We tested the effect of image super-resolution in surveillance FR on QMUL-SurvFace.
From Table \ref{table:SRFR_open}, we have two observations:
\noindent{(1)} 
Surprisingly, exploiting super-resolution algorithms often bring 
slightly {\em negative} effect to surveillance FR.
The plausible reasons are threefold.
The {\em first} is due to that the MSE loss (Eqn. \eqref{eq:MSE})
of super-resolution models
is not a perceptual measurement, but
a low-level pixel-wise metric.
The {\em second} is that 
the training data for super-resolution are
CASIA web faces with a domain gap against
QMUL-SurvFace images
(Fig. \ref{fig:FR_datasets}).
%
The {\em third} is the negative effect of
artefacts generated in super-resolving
(Fig. \ref{fig:SR_visualisation}).
A slight exception case  (similar as in Sec. \ref{sec:eval_LR_FR})
is VggFace due to the need for higher-resolution inputs therefore
somewhat preference towards super-resolution.
Besides, VggFace is the weakest in performance.
\noindent{(2)} Joint training of FR and super-resolution is 
not necessarily superior than independent training.
This suggests that it is non-trivial 
to effectively propagate the FR discrimination capability
into the learning of super-resolution.
Therefore, it is worth further in-depth investigation 
on how to integrate a super-resolution ability 
with surveillance FR.

\begin{table} [h] 
	\centering
	\setlength{\extrarowheight}{0.1mm}
	\setlength{\tabcolsep}{0.25cm} 
	\caption{
		Effect of super-resolution on
		down-sampled {\em MegaFace2} data. 
		{Protocol}: Open-Set.
	}
	\vskip -0.2cm
	\label{table:SRFR_mega_open}
	\begin{tabular}{l||cccc|c} 
\hline
		\multirow{2}{*}{Metrics}
		& \multicolumn{4}{c|}{TPIR$20$(\%)@FPIR} & \multirow{2}{*}{AUC (\%)}
		\\
		\cline{2-5} 
		& 30\% & 20\% & 10\% & 1\% & 
		\\
		\hline 
		\hline
		\backslashbox{SR \kern -1em}{FR}
		& \multicolumn{5}{c}{CentreFace}  
		\\
		\hline
		No SR  	
		& {39.9} & {28.0} & {14.0} & {5.8} 
		& {46.0}
		\\
		\hline 
		SRCNN
		& {26.6} & {19.2} & {10.0} & {5.0} 
		& {36.5}
		\\
		FSRCNN
		& {26.3} & {19.5} & {11.0} & {5.2} 
		& {35.3}
		\\
		VDSR
		& \bf 40.0 & \bf {28.3} & \bf {14.1} & \bf {6.0} 
		& \bf {47.5}
		\\
		\hline 
	\end{tabular}
\end{table}

\begin{figure} [h] 
	\centering
	\begin{subfigure}[t]{0.48\textwidth}
		\includegraphics[width=1\textwidth]{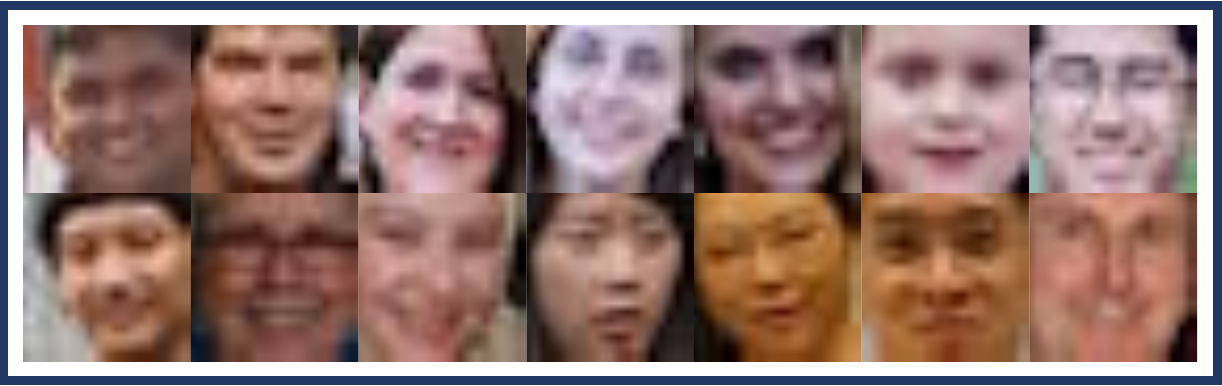}
		\caption{Simulated low-resolution MegaFace2 images.}
	\end{subfigure}
	\begin{subfigure}[t]{0.48\textwidth}
		\includegraphics[width=1\textwidth]{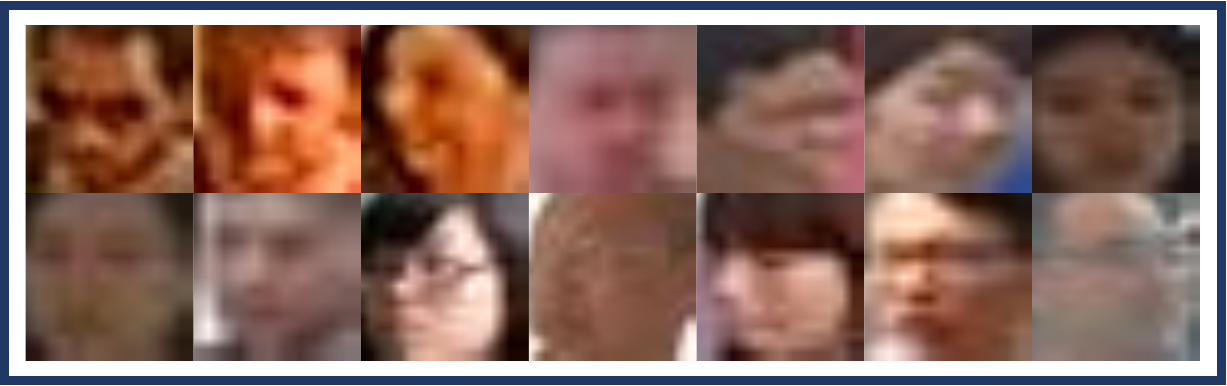}
		\caption{Native low-resolution QMUL-SurvFace.}
	\end{subfigure}
	\vskip -0.2cm
	\caption{
		Simulated {\em vs}
		native low-resolution faces.
	}
	\label{fig:MF2_vs_surv}
\end{figure}

\begin{figure} [h] 
	\centering
	\includegraphics[width=0.5\textwidth]{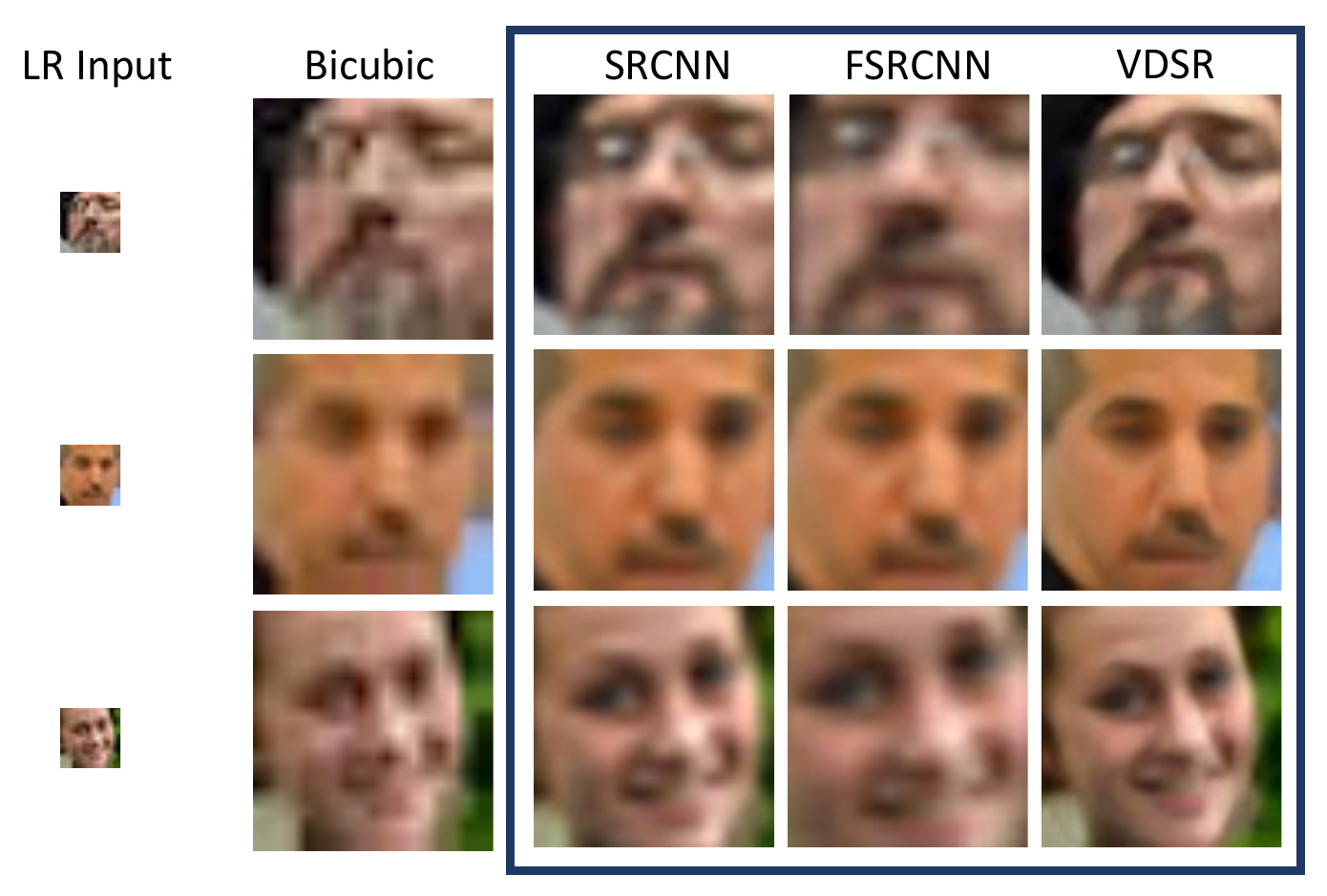}
	\vskip -0.2cm
	\caption{Super-resolved MegaFace2 images.
		{CentreFace} is used jointly with the 
		super-resolution models.
	}
	\label{fig:SR_visualisation_mega}
\end{figure}

\vspace{0.1cm}
\noindent {\bf (II) SurvFace {\em vs} WebFace.}
We evaluated the effect of super-resolution
on down-sampled web FR as a comparison 
to surveillance FR.
This mitigates the domain gap between training and test data
as encountered on QMUL-SurvFace.

\vspace{0.1cm}
\noindent {\em Setting.}
We constructed a low-resolution web FR dataset 
with a similar setting 
as QMUL-SurvFace (Table \ref{table:survface_datasplit})
by sampling MegaFace2 \citep{nech2017level}.
MegaFace2 was selected since it contains non-celebrity people 
thus ensuring ID non-overlap with the training data 
CASIA. 
We down-sampled selected MegaFace2 images  
to the mean QMUL-SurvFace size 24$\times$20
(Fig. \ref{fig:MF2_vs_surv}),
%
%
%
%
and built an open-set test setting 
with 3,000 gallery IDs (51,949 images)
and 10,254 probe IDs (176,990 images) (Table \ref{table:survface_veri}).
We further randomly sampled an ID-disjoint training set with
81,355 images of 5,319 IDs. 
In doing so, we created a like-for-like setting with low-resolution MegaFace2 
against QMUL-SurvFace.
We adopted the most effective {joint training} strategy.

\vspace{0.1cm}
\noindent {\em Results.} 
Table \ref{table:SRFR_mega_open} shows that
super-resolution at most brings very marginal gain
to low-resolution FR performance,
suggesting that 
contemporary techniques
are still far from satisfactory  
in synthesising FR discriminative fidelity 
(Fig. \ref{fig:SR_visualisation_mega}). 
In comparison, the FR performance assisted by VDSR in particular on simulated low-resolution web faces
is better than on surveillance images,
similarly reflected in super-resolved images
(Fig. \ref{fig:SR_visualisation_mega} vs Fig. \ref{fig:SR_visualisation}).
This indicates that low-resolution surveillance FR 
is more challenging due to the lack 
of high-resolution surveillance data.

\begin{table}[!htbp]
	\centering
	\setlength{\extrarowheight}{0.1mm}
	\setlength{\tabcolsep}{0.2cm} 
	\caption{
		Effect of super-resolution (SR) in face verification on
		{\em  QMUL-SurvFace}.
	}
	\vskip -0.2cm
	\label{table:surv_results_ver}
		\begin{tabular}{l|cccc|c}
		\hline 
		\multirow{2}{*}{Metrics}
		& \multicolumn{4}{c|}{TAR(\%)@FAR} & \multirow{2}{*}{AUC (\%)} \\
		\cline{2-5}
		& 30\% & 10\% & 1\% & 0.1\% & \\
		\hline \hline
		\backslashbox{SR \kern 0em}{FR}
		& \multicolumn{5}{c}{CentreFace \citep{wen2016discriminative}} \\
		\hline
		No SR 
		& {\bf 95.2} & {\bf 86.0} &  {\bf 53.3} & {\bf 26.8} & {\bf 94.8} \\
		\hline
		SRCNN 
		& 94.0 & 82.0 & 48.1 & 25.0 & 93.3
		\\
		FSRCNN 
		& 93.8 & 81.6 & 46.0 & 22.0 & 93.0
		\\
		VDSR 
		& 95.0 & 84.0 & 50.0 & 26.0 & 94.5
		\\
		\hline
	\end{tabular}
\end{table}

\subsubsection{Face Verification Evaluation}
We evaluated
the effect of super-resolution for pairwise face verification on 
QMUL-SurvFace.
%
We used the jointly trained CentreFace model.
%
Table \ref{table:surv_results_ver} shows that 
all methods decrease the performance slightly,
consistent with the face identification test above (Table \ref{table:SRFR_open}). 
%
Moreover, the verification result is largely unsatisfactory 
at strict false alarm rates, e.g. FAR=0.1\%.
Yet, this is very desired for real-world applications
because FAR is closely concerned with the system usability.
This indicates that there exists a large room for image super-resolution and
hallucination
towards improving face verification in surveillance facial images.


\section{Discussion and Conclusion}
In this work, we have presented a large surveillance face recognition benchmark 
called {\em QMUL-SurvFace}, 
including a real-world surveillance face dataset with native
low-resolution facial images,
extensive benchmarking experimental results,
in-depth discussions and analysis.
In contrast to existing FR challenges on which the model performances have
saturated, this challenge shows that state-of-the-art algorithms 
remain unsatisfactory 
in handling poor quality surveillance face images.
%
In concluding remarks, we discuss 
a number of research directions which we consider to be
worthwhile investigating in the future research.

\vspace{0.1cm}
\noindent\textbf{Transfer Learning.}
From the benchmarking results (Table \ref{table:SRFR_open}), 
it is shown that transferring the knowledge of auxiliary web facial data
is beneficial to boost the surveillance FR performance.
However, more research is needed to achieve more effective transfer.
Given the domain discrepancy between 
surveillance and web face images,
it is important to consider domain adaptation in training 
\citep{pan2010survey,tzeng2015simultaneous,saenko2010adapting,ganin2015unsupervised,ghifary2016deep}. 
%
Among existing solutions,
style transfer \citep{gatys2016image,li2016precomputed,li2017diversified,zhu2017unpaired}
is a straightforward approach. 
The idea is to transform the source images with 
the target domain style 
so that
the labelled source data 
can be exploited to train supervised FR models.
Whilst solving the style transfer problem is inherently challenging, 
it promises worthwhile potential for surveillance FR.

\vspace{0.1cm}
\noindent\textbf{Resolution Restoration.}
As mentioned earlier, the low-resolution nature hinders the surveillance FR performance.
While image super-resolution (SR) is one natural solution,
our evaluations show that the current algorithms
are not effective. 
Two main reasons are: 
(1) We have no access to native high-resolution surveillance faces
required in training SR models
\citep{yang2014single}.
(2) It is difficult to transfer the SR model
learned on web data due to the domain gap
\citep{pan2010survey}. 
Although SR techniques have been adopted 
to improve low-resolution FR \citep{wang2014low},
they rely on hand-crafted feature representations with 
limited test on small simulated data.
It remains unclear how effective SR methods
are for native surveillance FR.
%

\vspace{0.1cm}
\noindent\textbf{Semantic Attributes.}
Face attributes serve as a mid-level representation
to boost FR in high-resolution facial images
\citep{kumar2009attribute,berg2013poof,manyam2011two,song2014exploiting}.
In general, it is challenging to detect attributes in face due to 
complex covariates in pose, expression, lighting, occlusion, and background
\citep{farhadi2009describing,parikh2011relative},
besides imbalanced class distributions
\citep{dong2018imbalanced,dong2017class,huang2016learning}.
And this is even harder given poor quality surveillance images.
This direction becomes more plausible due to 
the emergence of large attribute datasets \citep{liu2015faceattributes}.
%
We anticipate that attributes will play an important role 
in the development of surveillance FR techniques.

\begin{figure} [h] 
	\centering
	\begin{subfigure}[t]{0.235\textwidth}
		\includegraphics[width=1\textwidth]{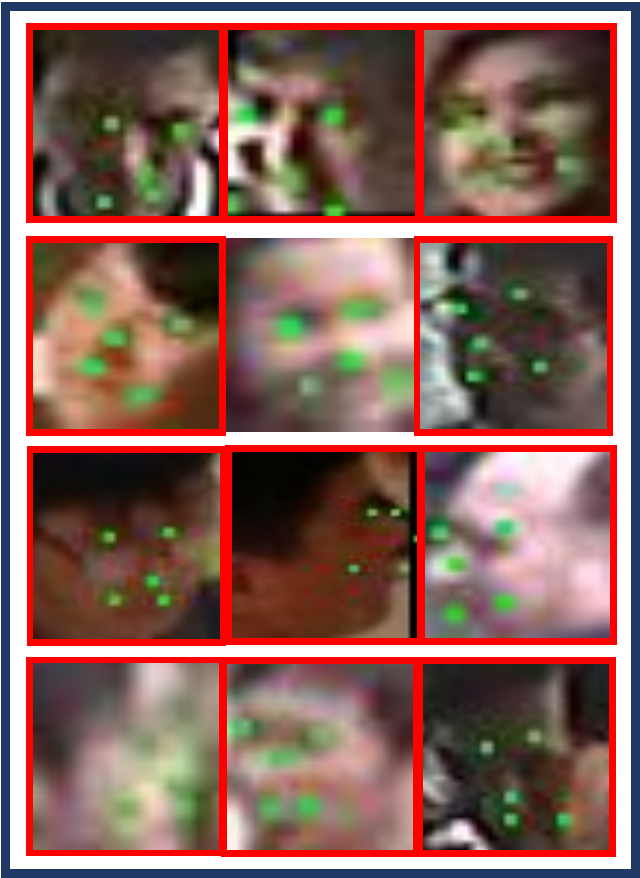}
		\caption{QMUL-SurvFace}
	\end{subfigure}
	\begin{subfigure}[t]{0.2316\textwidth}
		\includegraphics[width=1\textwidth]{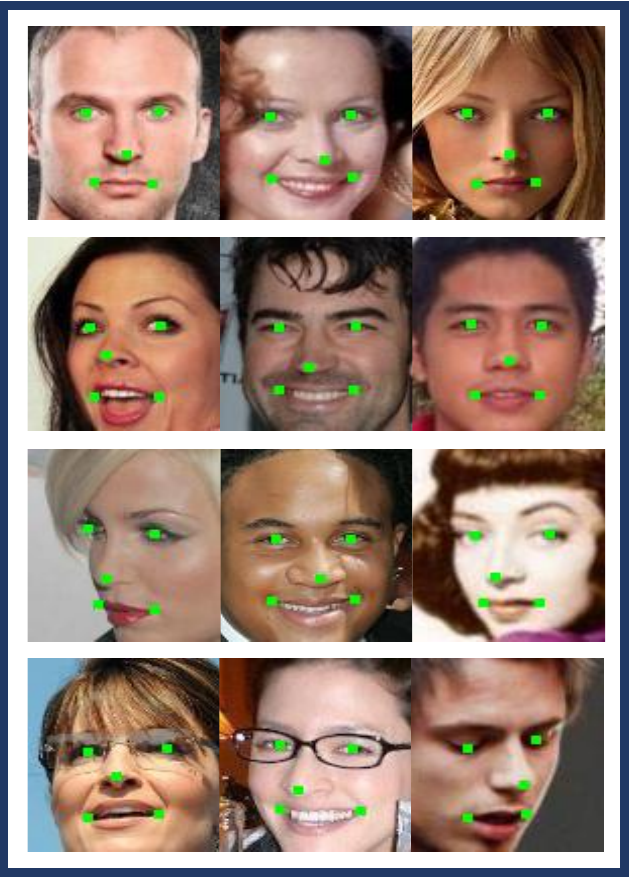}
		\caption{CASIA web faces}
	\end{subfigure}
	\vskip -0.2cm
	\caption{
		Facial landmark detection \citep{ZhangSPL16} on
		randomly sampled 
		(a) QMUL-SurvFace
		and (b) CASIA web faces.
		Landmarks include left/right eye centres, nose tip, and left/right mouth corners, 
		denoted by green dots. 
		Failure cases
		are indicated by red box. 
	}
	\label{fig:alignment}
\end{figure}

\vspace{0.1cm}
\noindent\textbf{Face Alignment.}
Face alignment by landmark detection is an indispensable preprocessing
in FR \citep{chen2012bayesian,wen2016discriminative,ZhangSPL16}.
Despite the great progress
\citep{zhu2015face,zhang2014facial,cao2014face},
aligning face remains a formidable challenge especially in surveillance images
(Fig. \ref{fig:alignment}).
%
Similar to SR and attribute detection,
this task suffers the domain shift.
%
One idea is hence to construct a large 
surveillance face landmark dataset.
%
Integrating landmark detection with SR is also 
an interesting topic.

\vspace{0.1cm}
\noindent\textbf{Contextual Constraints.}
Given incomplete and noisy observation in surveillance facial data,
it is important to discover and model context information as 
extra constraint. 
In social environment, people often travel in groups.
The group structure provides useful information (social force)
for model inference 
\citep{gallagher2009understanding,zheng2009associating,helbing1995social,hughes2002continuum}.

\vspace{0.1cm}
\noindent\textbf{Open-Set Recognition.}
Surveillance FR is an open-set recognition problem \citep{bendale2015towards,bendale2016towards}.
In reality, most probes are non-target persons. 
It is hence beneficial that the model learn to construct a decision boundary for the target people \citep{zheng2016towards}.
Whilst open-set recognition techniques evolve independently,
we expect more future attempts at jointly solving the two problems.

\vspace{0.1cm}
\noindent\textbf{Zero-Shot Learning.}
FR is about zero-shot learning (ZSL) 
with no training samples of test classes available
\citep{fu2017recent}.
In the literature, the focus of ZSL is knowledge transfer across seen and unseen classes
via intermediate representations such as attributes \citep{fu2014learning} 
and word vectors \citep{frome2013devise}.
All classes are assigned with fixed representations.
%
On the contrary, FR does not use such bridging information but
learns an ID-sensitive representation from training classes.
%
In this sense, FR is more generalised. 

\vspace{0.1cm}
\noindent\textbf{Imagery Data Scalability.}
Compared to existing web FR benchmarks \citep{kemelmacher2016megaface,nech2017level,liu2015faceattributes,yi2014learning},
the QMUL-SurvFace is smaller in scale.
An important future effort is to 
expand this challenge for more effective 
model training and further scaling up the open-set test.
%

%
\vspace{0.1cm}
\noindent\textbf{Final Remark.}
Given that 
FR performance has largely saturated on existing web face challenges,
this work presents timely a more challenging benchmark QMUL-SurvFace 
for further stimulating innovative algorithms.
This benchmark calls for more research efforts 
for the under-studied and crucial surveillance FR problem.

\section*{Acknowledgements}
This work was partially supported by
the Royal Society Newton Advanced Fellowship
Programme (NA150459), 
Innovate UK Industrial Challenge Project on Developing and Commercialising Intelligent Video Analytics Solutions for Public Safety (98111-571149),
Vision Semantics Ltd,
and SeeQuestor Ltd.

\bibliographystyle{spbasic}      
\bibliography{survFR}   


\end{document}